\documentclass[a4paper,fleqn]{cas-dc}

\usepackage[authoryear,longnamesfirst]{natbib}

\usepackage{tabularx}
\usepackage{textcomp}
\usepackage{algorithm,algpseudocode}
\usepackage{mathrsfs}
\usepackage{placeins}
\usepackage{xurl}

\begin{document}
\let\WriteBookmarks\relax
\let\printorcid\relax 

\shorttitle{Evidence-Grounded Iterative Reasoning for DCLDs}
\shortauthors{Li et al.}

\title[mode = title]{Evidence-Grounded Iterative Reasoning for the Diagnosis of Diffuse Cystic Lung Diseases}

\author[1,2]{Haoqing Li}
\ead{li_haoqing@mail.ustc.edu.cn}

\author[1]{Jun Shi}
\cormark[1]
\ead{shijun18@ustc.edu.cn}

\author[1]{Tou Li}
\author[3]{Xianmeng Chen}
\author[4]{Rui Wang}
\author[6]{Yating Peng}
\author[5]{Qiwei Jia}
\author[6]{Li Gao}
\author[6]{Qianyi Chen}
\author[1]{Hong An}
\author[6]{Ruoyun Ouyang}
\author[3]{Xiaowen Hu}

\affiliation[1]{organization={School of Computer Science and Technology, University of Science and Technology of China},
                  city={Hefei},
                  country={China}}
\affiliation[2]{organization={Laoshan Laboratory},
                  city={Qingdao},
                  country={China}}
\affiliation[3]{organization={Department of Pulmonary and Critical Care Medicine, The First Affiliated Hospital of USTC, Division of Life Sciences and Medicine, University of Science and Technology of China},
                  city={Hefei},
                  country={China}}
\affiliation[4]{organization={Department of Respiratory Medicine, No. 901 Hospital of the Chinese People's Liberation Army Logistic Support Force},
                  city={Hefei},
                  country={China}}
\affiliation[5]{organization={School of the Gifted Young, University of Science and Technology of China},
                  city={Hefei},
                  country={China}}
\affiliation[6]{organization={Department of Pulmonary and Critical Care Medicine, The Second Xiangya Hospital, Central South University},
                  city={Changsha},
                  country={China}}

\cortext[1]{Corresponding author: Jun Shi.}

\begin{abstract}
Overlapping imaging manifestations complicate the differential diagnosis of Diffuse Cystic Lung Diseases (DCLDs), while limited annotated data constrain supervised model development. Multimodal Large Language Models (MLLMs) can integrate imaging and clinical information, but their diagnostic reasoning remains limited by incomplete volumetric perception and insufficient grounding in disease-specific evidence. We propose DCLDs-EIR, an evidence-grounded iterative reasoning framework coupling diagnostic hypothesis refinement with multimodal evidence acquisition and verification. Multiview renderings of predicted lung and cyst segmentations characterize global cyst distributions and initialize the differential diagnosis. Competing hypotheses guide axial CT inspection to identify local findings that, together with clinical context, inform retrieval from a disease-specific knowledge graph. Accumulated patient observations and retrieved knowledge are jointly assessed for diagnostic support and refutation to revise the diagnostic state. Updated hypotheses and unresolved distinctions guide subsequent inspection, forming a feedback loop between evidence acquisition and hypothesis verification. Multicenter evaluation demonstrates that DCLDs-EIR consistently improves diagnostic accuracy over baselines and enables traceable differential diagnosis through explicit links between diagnostic hypotheses, patient observations, and disease-specific knowledge.
\end{abstract}

\begin{graphicalabstract}
\includegraphics[width=\textwidth]{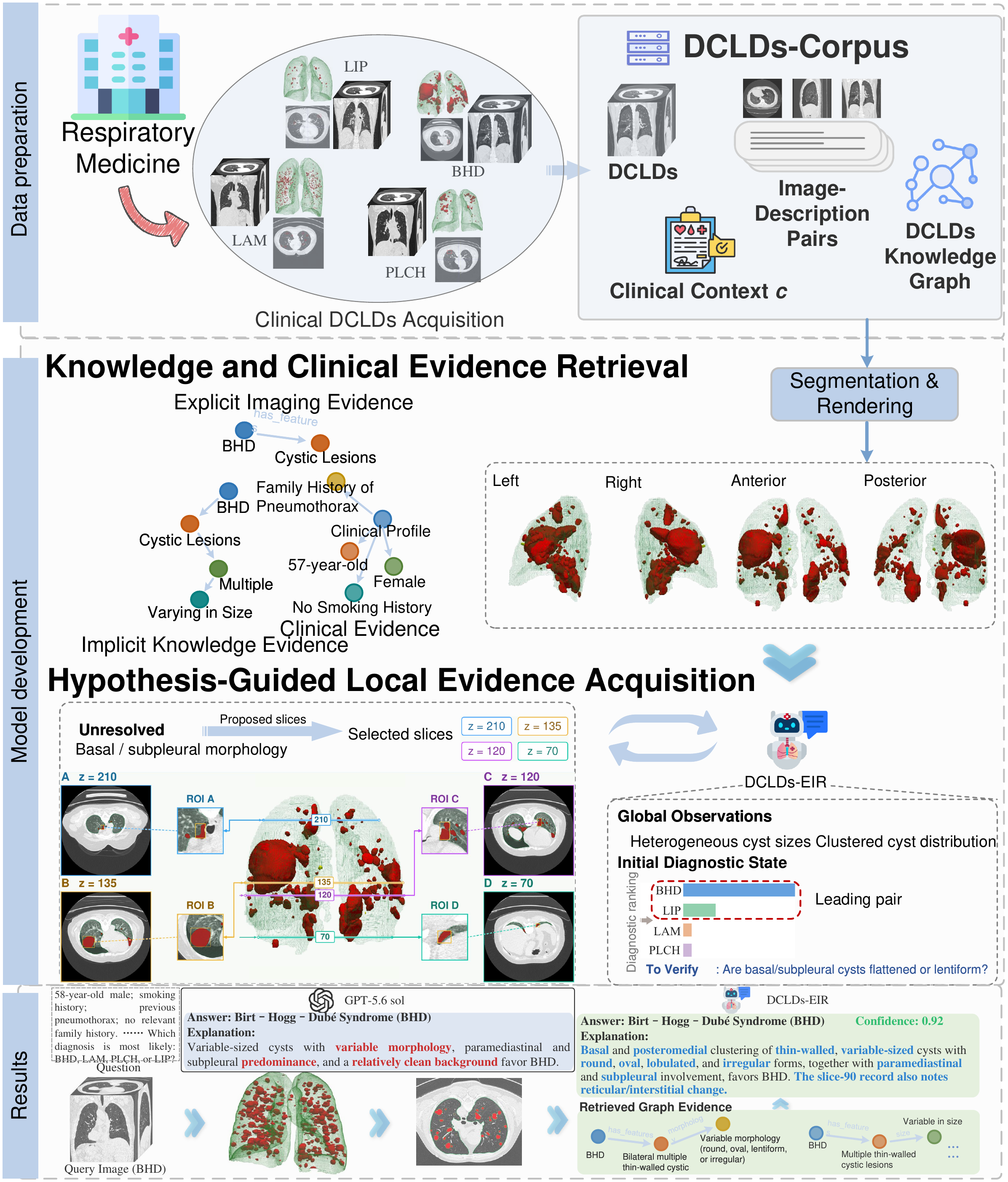}
\end{graphicalabstract}

\begin{highlights}
\item Global lung renderings guide targeted inspection of computed tomography slices.
\item Iterative retrieval connects disease knowledge with patient-specific findings.
\item Evidence verification assesses support and refutation for competing diagnoses.
\item External accuracy reached 90.00\% in an independent examination cohort.
\end{highlights}


\maketitle

\section{Introduction}
\label{sec:introduction}

\begin{figure*}[pos=!tbp]
\centering
\includegraphics[width=0.85\textwidth]{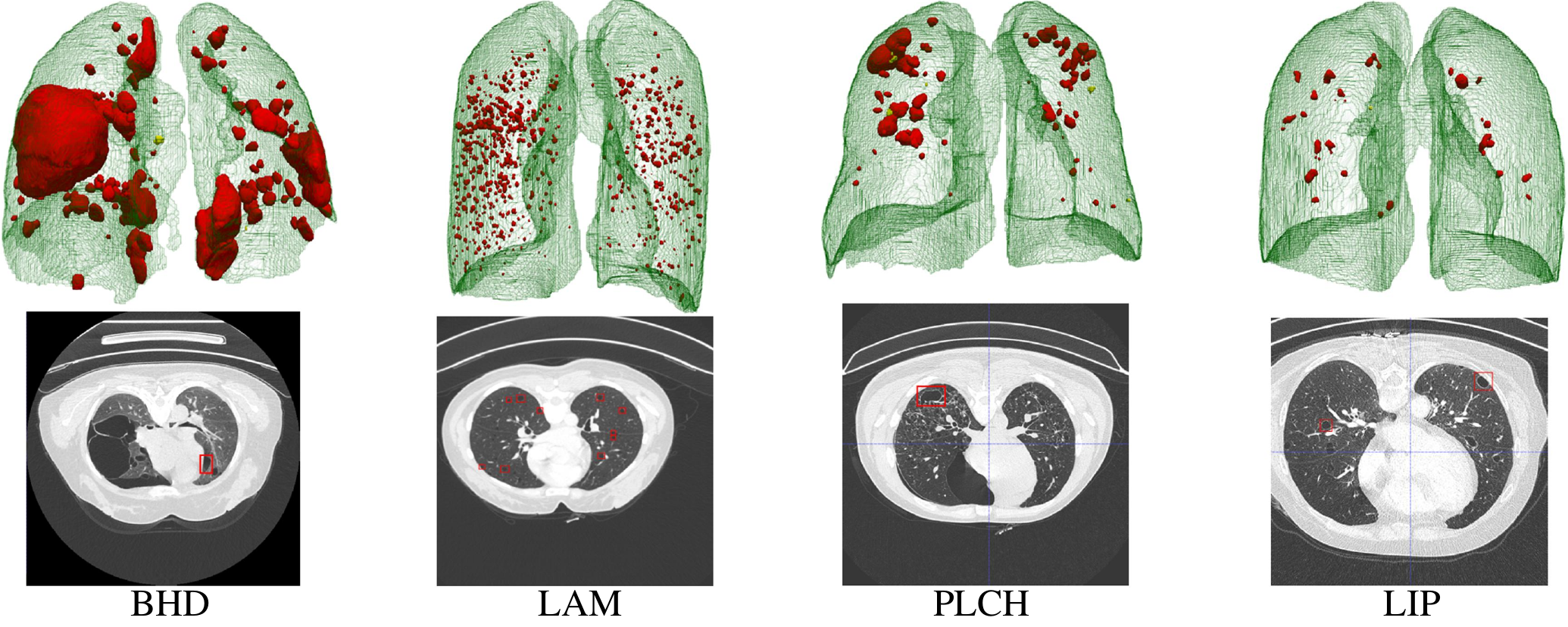}
\caption{Representative 3D renderings and CT slices of DCLDs subtypes.}
\label{fig:dcld_examples}
\end{figure*}

Diffuse cystic lung diseases (DCLDs) comprise rare pulmonary disorders with overlapping manifestations on computed tomography (CT)~\citep{p1577_r01,p1577_r02}. Differentiating Birt--Hogg--Dub\'e syndrome (BHD), lymphangioleiomyomatosis (LAM), pulmonary Langerhans cell histiocytosis (PLCH), and lymphocytic interstitial pneumonia (LIP) requires joint assessment of cyst morphology, whole-lung distribution, associated parenchymal abnormalities, and clinical history~\citep{p1577_r03,p1577_r05}. Representative appearances are shown in Fig.~\ref{fig:dcld_examples}. The combination of overlapping appearances and limited annotated data complicates the development of supervised models that capture these context-dependent diagnostic distinctions.

Advances in medical question answering~\citep{p323_r26} motivate the use of language models for diagnostic reasoning. Multimodal large language models (MLLMs) provide a common interface for integrating images, clinical text, and external knowledge, offering opportunities for rare-disease diagnosis without task-specific fine-tuning. However, radiological evaluations reveal limitations in image interpretation~\citep{p1577_r19}. For DCLDs, effective reasoning requires both sufficient patient observations across the CT volume and disease-specific knowledge to assess their relevance to competing diagnoses.

The first challenge is acquiring sufficient patient evidence from volumetric images. Whole-lung cyst distribution provides global anatomical context, whereas original axial CT slices reveal local wall morphology and associated parenchymal findings. A small set of representative slices may omit findings needed for differential diagnosis, while extensive slice processing increases inference cost. Global renderings can summarize lesion distribution and guide inspection but cannot independently establish fine morphological details. Moreover, the diagnostic value of a local finding depends on which hypotheses remain plausible. Efficient volumetric reasoning therefore requires selecting slices for inspection according to both spatial coverage and unresolved distinctions between competing diagnoses.

The second challenge is determining how disease-specific knowledge supports or refutes the current hypotheses. Medical language-model benchmarks document factual errors~\citep{p1577_r17}. Retrieval-augmented generation (RAG) supplements model knowledge with external information~\citep{p323_r07}, but irrelevant retrieved context can impair reasoning~\citep{p323_r32}. Even a correct disease association does not establish that the corresponding finding is present in the patient. An unassessed feature must also be distinguished from an observed negative finding. Diagnostic support or refutation must therefore be assessed by comparing retrieved associations with observed imaging findings and clinical information. This assessment must also consider findings that favor alternative diagnoses or contradict the leading hypothesis. When a relevant observation is missing, additional retrieval or repeated reasoning alone does not supply the missing imaging evidence.

Our previous studies address these challenges from complementary perspectives. The spatial self-verification framework~\citep{jia_selfverification_miccai} combines global spatial perception (GSP) with targeted axial CT inspection to refine diagnostic hypotheses, while interpretation of the acquired findings relies on the MLLM's existing disease knowledge. DCLDs-RAG~\citep{li_dcldsrag_miccai} supplements patient inputs with retrieved visual references and structured knowledge and clinical evidence. Its retrieval augments a fixed set of patient observations and is not coupled to further CT inspection. These complementary limitations motivate feedback between knowledge retrieval and patient evidence acquisition. Disease knowledge can indicate which observations would distinguish competing hypotheses, while new patient findings can refine retrieval and determine whether the retrieved associations are diagnostically relevant.

We propose DCLDs-EIR, an evidence-grounded iterative reasoning framework that couples diagnostic hypothesis refinement with multimodal evidence acquisition and verification. Multiview GSP summarizes whole-lung cyst distribution from predicted lung and cyst segmentations and initializes the differential diagnosis. The leading hypotheses then guide axial CT inspection, and newly acquired observations together with clinical information condition retrieval from a disease-specific knowledge graph. Evidence-grounded hypothesis verification assesses retrieved associations against accumulated patient findings and updates the diagnostic ranking by considering supporting and refuting evidence. Updated hypotheses and unresolved distinctions guide subsequent inspection, allowing each round to address the evidence gaps identified in the preceding assessment. Accumulated evidence retains explicit links to its sources, while predefined evidence and ranking-stability criteria govern diagnostic acceptance.

The main contributions are summarized below.
\begin{enumerate}
\item We introduce a hypothesis-guided framework that couples global spatial perception, targeted CT inspection, and knowledge and clinical evidence retrieval across verification rounds. Acquired observations refine retrieval, while updated hypotheses and retrieved knowledge guide subsequent CT inspection.

\item We develop evidence-grounded hypothesis verification that assesses support and refutation for competing diagnoses using accumulated evidence. Explicit links between diagnostic hypotheses, patient observations, and retrieved knowledge provide a traceable basis for diagnostic revision and acceptance.

\item We evaluate DCLDs-EIR across four disease categories on internal and external cohorts without task-specific MLLM fine-tuning. With GPT-5.6 Sol, the framework achieves 90.00\% accuracy and 89.05\% macro-F1 on the independent external cohort.
\end{enumerate}

\section{Related Work}
\label{sec:related_work}

\subsection{MLLMs for Medical Diagnosis}

Medical MLLMs adapt image--language representations to clinical interpretation through biomedical alignment and instruction tuning. LLaVA-Med~\citep{p1577_r07} learns from biomedical figure--caption pairs and visual instructions, whereas HuatuoGPT-Vision~\citep{huatuo} uses refined medical visual question--answer data. These approaches support biomedical dialogue and question answering, complementing general-purpose GPT, Gemini, Qwen, and Claude models~\citep{gpt56sol,gemini25pro,qwenvlmax,claude45}.

Volumetric interpretation additionally requires integrating spatial context across CT slices. RadFM and M3D-LaMed extend image--language modeling to 3D data~\citep{p1577_r09,p1577_r08}, while CTInstruct combines multi-granular pretraining with instruction tuning for multiple CT tasks~\citep{ctinstruct}. Med3DVLM~\citep{med3dvlm} emphasizes efficient volumetric encoding and image--text alignment, whereas Med-2E3~\citep{med2e3} uses text-guided slice weighting to integrate 2D and 3D features. Reg2RG~\citep{chen2025reg2rg} further connects global and regional CT features to report descriptions through region--report alignment. Recent medical image analysis studies have also investigated disease-aware representations and lesion analysis in lung CT: DrasCLR learns disease-related and anatomy-specific representations from 3D lung CT~\citep{mia_yu2024drasclr}, pathological-prior-guided feature synthesis has been explored for CT-based lung cancer subtype classification~\citep{mia_zhu2024lungct}, and unsupervised anomaly-mask synthesis has been used for pulmonary disease localization~\citep{mia_hao2025pulmonary}. These studies advance volumetric representation and pulmonary image analysis, while DCLDs diagnosis additionally requires selecting patient observations that distinguish competing hypotheses and verifying their diagnostic relevance using disease-specific knowledge.

\subsection{Graph Retrieval and Iterative Reasoning}

Retrieval augmentation supports medical question answering~\citep{xiong2024mirage}. GRAG~\citep{p323_r12} exploits subgraph text and topology, while Medical Graph RAG~\citep{p323_r29} links user documents to medical references. STREAM~\citep{yang2025stream} retrieves anatomical entities for chest X-ray report generation, whereas CKRA~\citep{yang2026ckra} combines question-relevant knowledge selection with knowledge--image contrastive learning for medical visual question answering. Related medical image analysis work has explored knowledge-guided multimodal generation and graph-based clinical reasoning: learned knowledge bases and image--report alignment have been used for chest X-ray report generation~\citep{mia_yang2023kb}, multimodal relationship graphs have supported interpretable medical visual question answering~\citep{mia_hu2024graphvqa}, and topic graphs have been incorporated into topic-guided radiology report generation~\citep{mia_li2025topic}.

Intermediate reasoning and self-verification~\citep{p1577_r20,p1577_r22} can be integrated with retrieval~\citep{trivedi2023ircot,asai2024selfrag}. Recent medical VQA studies have further examined interpretability and modality-specific bias: UnICLAM uses contrastive representation learning with adversarial masking~\citep{mia_zhan2025uniclam}, while CIMB-MVQA applies causal intervention to reduce visual and language confounders~\citep{mia_liu2026cimb}. \citet{xu2025causalmedvqa} address cross-modal bias in medical visual question answering through causal modeling and LLM prompting, while Med-R1~\citep{lai2026medr1} uses reinforcement learning to improve medical reasoning across imaging modalities.

DCLDs-RAG~\citep{li_dcldsrag_miccai} augments diagnostic reasoning through knowledge--clinical retrieval conditioned on fixed patient observations. Spatial self-verification~\citep{jia_selfverification_miccai} iteratively acquires additional CT observations, with diagnostic interpretation constrained by the MLLM's rare-disease knowledge. Our DCLDs-EIR integrates these complementary capabilities through hypothesis-guided coordination and knowledge--clinical retrieval. Acquired observations inform retrieval and hypothesis verification, while retrieved knowledge and updated hypotheses guide subsequent CT inspection.

\begin{figure*}[pos=!tbp]
\centering
\includegraphics[width=\textwidth]{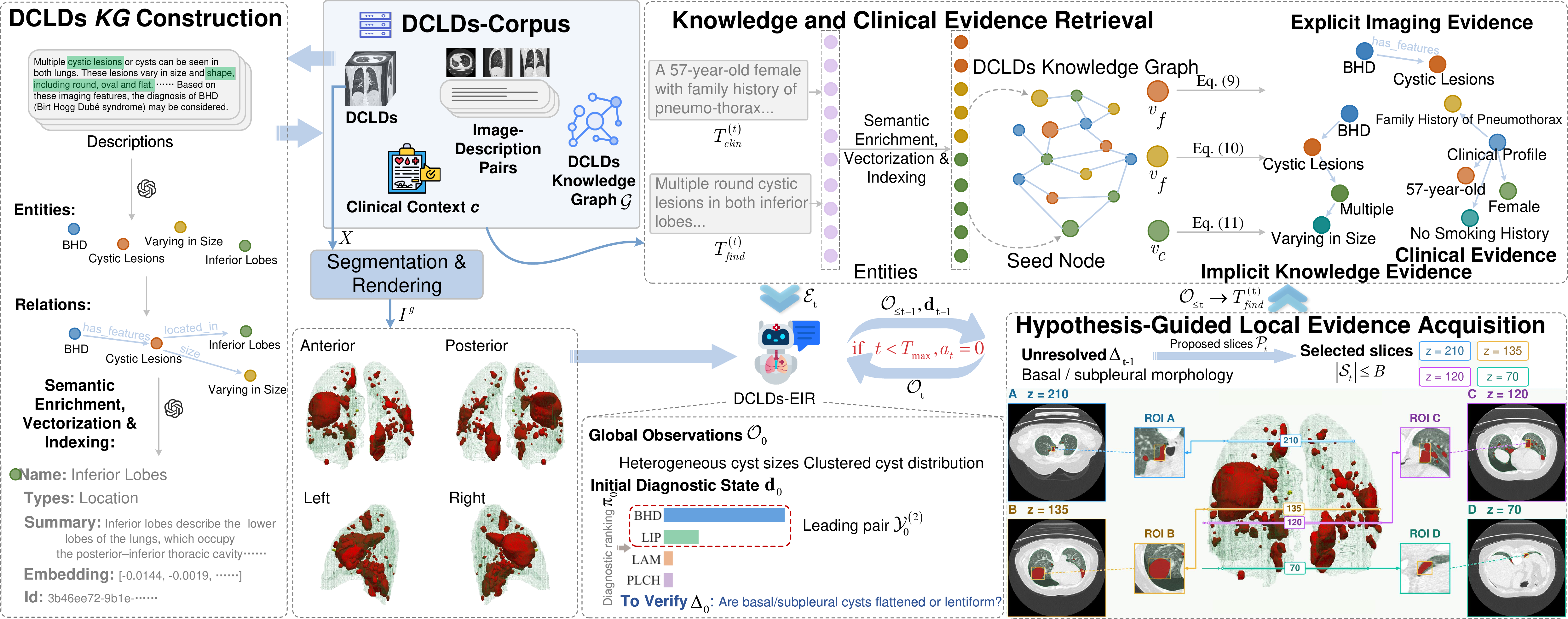}
\caption{Overview of DCLDs-EIR for evidence-grounded iterative diagnosis. A disease-specific knowledge graph is constructed from imaging descriptions. Multiview lung renderings capture global cyst distributions to initialize the differential diagnosis. Unresolved diagnostic differences guide targeted CT inspection, and updated findings with clinical context guide graph retrieval. Accumulated patient observations and retrieved knowledge are integrated to verify competing hypotheses, revise the diagnostic ranking, and guide subsequent inspection.}
\label{fig:main}
\end{figure*}

\section{Methods}
\label{sec:methods}

\subsection{Problem Formulation and Framework Overview}
\label{sec:method_overview}

Given a CT volume $X\in\mathbb R^{D\times H\times W}$ and clinical context $\mathbf c$, DCLDs-EIR predicts a diagnosis in $\mathcal Y=\{\mathrm{BHD},\mathrm{LAM},\mathrm{PLCH},\mathrm{LIP}\}$ together with supporting evidence. Here, $D$, $H$, and $W$ denote volume depth, height, and width, and $\mathbf c$ comprises available demographics, clinical history, and accompanying imaging findings.

As illustrated in Fig.~\ref{fig:main}, DCLDs-EIR coordinates spatial perception and disease knowledge through iterative verification of competing diagnoses. Multiview global spatial perception characterizes whole-lung cyst distributions to establish an initial differential diagnosis~\citep{jia_selfverification_miccai}, which guides local CT inspection toward findings that distinguish the leading hypotheses. Knowledge and clinical evidence retrieval uses these observations and the clinical context to identify relevant associations in a disease-specific knowledge graph $\mathcal G$~\citep{li_dcldsrag_miccai}, supplementing the MLLM's knowledge of rare diseases. Evidence-grounded hypothesis verification evaluates these associations against patient findings to determine diagnostic support and counterevidence, revise the ranking, and identify unresolved distinctions.

The revised hypotheses, unresolved distinctions, and retrieved knowledge guide subsequent CT inspection, while newly acquired observations refine retrieval and diagnostic assessment. This coupling makes evidence acquisition responsive to the current differential diagnosis and grounds hypothesis revision in accumulated patient observations and disease knowledge. Diagnostic acceptance is governed by evidence sufficiency, unresolved conflicts, and ranking stability, with termination at the round limit distinguished in the output. Algorithm~\ref{alg:dclds_rag} summarizes the inference procedure.

\subsection{Multiview Global Spatial Perception}
\label{sec:method_gsp}

CT volumes and predicted masks share a canonical coordinate system and axial indexing. An nnU-Net model $\mathcal F_\phi$~\citep{p1577_r06}, with fixed parameters $\phi$, produces binary lung and cyst masks $(M_{\mathrm{lung}},M_{\mathrm{cyst}})$, which are combined into a label map $L$:
\begin{equation}
\begin{aligned}
(M_{\mathrm{lung}},M_{\mathrm{cyst}})&=\mathcal F_{\phi}(X),\\
L&=\operatorname{Fuse}(M_{\mathrm{lung}},M_{\mathrm{cyst}}).
\end{aligned}
\label{eq:segmentation}
\end{equation}
Following~\citep{jia_selfverification_miccai}, marching cubes~\citep{p1577_r27} extracts lung and cyst surfaces, and PyVista~\citep{p1577_r32} renders cysts within transparent lungs. Anterior, posterior, left, and right views define $\mathcal V$, and their renderings are arranged into a multiview montage:
\begin{equation}
I^{\mathrm g}=\operatorname{Tile}\!\left(
\left\{\operatorname{Render}(L;\nu):\nu\in\mathcal V\right\}\right).
\label{eq:multiview}
\end{equation}
The composite represents global cyst distribution, relative size, and spatial density. We denote the MLLM by $\mathcal M_\theta$, with parameters $\theta$ fixed during inference. Initialization is indexed by $t=0$. Given $I^{\mathrm g}$, $\mathbf c$, and $D$, the MLLM returns the initial diagnostic state $\mathbf d_0$, global observations $\mathcal O_0$, and an ordered list $\mathcal P_1$ of proposed axial indices:
\begin{equation}
(\mathbf d_0,\mathcal O_0,\mathcal P_1)
=\mathcal M_{\theta}^{\mathrm{init}}(I^{\mathrm g},\mathbf c,D).
\label{eq:initial_state}
\end{equation}
Superscripts on $\mathcal M_\theta$ distinguish \textbf{init}ialization, \textbf{obs}ervation, and \textbf{ver}ification tasks performed by the same MLLM.

For initialization and subsequent rounds, the diagnostic state is $\mathbf d_t=(\pi_t,\mathbf s_t,\Delta_t,\mathcal B_t)$. Here, $\pi_t$ ranks the four diseases, $\mathbf s_t$ contains optional model scores, $\Delta_t$ records unresolved diagnostic distinctions and evidence conflicts, and $\mathcal B_t$ links candidate diagnoses to supporting or refuting records in the accumulated evidence $\mathcal L_t$. These records retain patient observations, retrieved knowledge, and their sources. With $\pi_t(r)$ denoting the disease at rank $r$, the leading pair is
\begin{equation}
\mathcal Y_t^{(2)}=\{\pi_t(1),\pi_t(2)\}.
\label{eq:leading_pair}
\end{equation}
The supplied context, global observations, and their sources initialize $\mathcal L_0$, with cumulative observations $\mathcal O_{\leq0}=\mathcal O_0$. Subsequent rounds $t=1,\ldots,T_{\max}$ examine local CT findings and update this state, where $T_{\max}$ is the round limit.

\subsection{Hypothesis-Guided Local Evidence Acquisition}
\label{sec:method_local}

Local inspection combines a fixed candidate pool with MLLM-proposed locations that evolve with the diagnostic state. After initialization, an ordered candidate pool $\mathcal I$ is constructed from the predicted masks and $\mathcal Y_0^{(2)}$, with indices in $\{0,\ldots,D-1\}$. Slice-wise cyst burden is
\begin{equation}
b(z)=
\frac{\sum_{i=0}^{H-1}\sum_{j=0}^{W-1}M_{\mathrm{cyst}}[z,i,j]}
{\max\!\left(\sum_{i=0}^{H-1}\sum_{j=0}^{W-1}M_{\mathrm{lung}}[z,i,j],1\right)}.
\label{eq:cyst_burden}
\end{equation}
High-burden slices are greedily selected with a minimum separation of eight slices. Additional candidates include the 15\%, 35\%, 50\%, 65\%, and 85\% volume-depth positions, burden maxima within three equal depth intervals, and the maximum-area slice of the largest three-dimensional cyst component. Disease-conditioned candidates emphasize inferior-half burden for BHD, volume-wide coverage and high burden for LAM, superior-region burden and nodule-related peaks for PLCH, and regional burden and background sampling for LIP. The resulting $\mathcal I$ remains fixed across rounds.

At round $t$, $\mathcal P_t$ contains the ordered MLLM proposals, and $\mathcal J_{t-1}$ contains previously inspected indices, with $\mathcal J_0=\varnothing$. The selected set is
\begin{equation}
\begin{aligned}
\mathcal S_t&=\operatorname{Select}
(\mathcal I,\mathcal P_t,\mathcal J_{t-1};B),\\
\mathcal S_t&\subseteq\{0,\ldots,D-1\},\qquad
|\mathcal S_t|\leq B,
\end{aligned}
\label{eq:slice_selection}
\end{equation}
where $B$ is the per-round slice budget, set to four in the main experiments. Selection prioritizes coverage-quantile candidates in $\mathcal I$, then valid MLLM proposals in their returned order, including those outside $\mathcal I$, and finally the remaining ranked candidates. Each round selects valid indices not in $\mathcal J_{t-1}$ with a minimum separation of eight slices within the round, and updates the inspection history as $\mathcal J_t=\mathcal J_{t-1}\cup\mathcal S_t$.

Lung-window transformation $\mathcal W$ produces $I_z=\mathcal W(X[z,:,:])$ and the local image set $I_t^{\mathrm{loc}}=\{I_z:z\in\mathcal S_t\}$. The MLLM examines these CT images without segmentation overlays:
\begin{equation}
\mathcal O_t=\mathcal M_{\theta}^{\mathrm{obs}}
\left(I_t^{\mathrm{loc}},\mathbf c,\mathbf d_{t-1}\right).
\label{eq:local_observation}
\end{equation}
The observation set $\mathcal O_t$ comprises imaging findings with corresponding anatomical locations and source slices. Cumulative observations and accumulated evidence are updated as $\mathcal O_{\leq t}=\operatorname{Append}(\mathcal O_{\leq t-1},\mathcal O_t)$ and $\mathcal L_t^{\mathrm{img}}=\operatorname{Append}(\mathcal L_{t-1},\mathcal O_t)$, respectively. The updated findings condition knowledge retrieval before revision of the diagnostic state.

\subsection{Knowledge and Clinical Evidence Retrieval}
\label{sec:method_retrieval}

The graph representation and retrieval scoring functions follow formulation~\citep{li_dcldsrag_miccai}. The knowledge graph is
\begin{equation}
\mathcal G=\left(E,R,\{\mathbf v_j\}_{j=1}^{|E|}\right),
\label{eq:knowledge_graph}
\end{equation}
where $E$ is the entity set, $R$ contains typed relation triples $(e_i,r,e_j)$, and $\mathbf v_j$ embeds entity $e_j\in E$. Entities represent diseases, imaging features and subfeatures, anatomical locations, patient groups, and clinical attributes. The graph and embeddings remain fixed during inference.

The clinical text $T_{\mathrm{clin}}$ is derived from $\mathbf c$, while $T_{\mathrm{find}}^{(t)}$ combines supplied imaging findings with $\mathcal O_{\leq t}$ and preserves their sources. Entity alignment yields the seed set $V_{\mathrm{seed}}^{(t)}\subseteq E$, including imaging-feature nodes $v_f$ and clinical nodes $v_c$.

\textit{Explicit imaging evidence.} Graph associations between imaging features and diseases are ranked by feature specificity:
\begin{equation}
S_v(v_f)=
\frac{\alpha}{\sqrt{\max\!\left(|\mathcal N_{\mathrm{dis}}(v_f)|,1\right)}},
\label{eq:specificity}
\end{equation}
where $\mathcal N_{\mathrm{dis}}(v_f)$ contains disease nodes adjacent to $v_f$, and $\alpha$ is a weight. Higher scores indicate greater disease specificity.

\textit{Implicit knowledge evidence.} Bounded traversal from $V_{\mathrm{seed}}^{(t)}$ identifies paths $p_k$ connecting imaging-derived entities to disease nodes, scored by
\begin{equation}
S_k(p_k)=
\frac{w_1\bar C_E+w_2\bar C_D+w_3\bar C_B+w_4\bar C_T}
{1+\gamma\,\operatorname{len}(p_k)}.
\label{eq:topology}
\end{equation}
Here, $\bar C_E$, $\bar C_D$, $\bar C_B$, and $\bar C_T$ are the mean eigenvector, degree, betweenness, and triangle-based node centralities along $p_k$. The weights $w_1,\ldots,w_4$ control their contributions, and $\gamma$ penalizes path length.

\textit{Clinical evidence.} Compatibility between clinical nodes $v_c$ and the patient is scored by
\begin{equation}
S_c(v_c)=\lambda_1\mathbb I(\mathrm{sex})
+\lambda_2\mathbb I(\mathrm{age})
+\lambda_3|H_{\mathrm{input}}\cap H_c|,
\label{eq:clinical_match}
\end{equation}
where the indicators assess sex and age compatibility with $v_c$. The sets $H_{\mathrm{input}}$ and $H_c$ contain history attributes from $T_{\mathrm{clin}}$ and $v_c$, respectively, and $\lambda_1,\lambda_2,\lambda_3$ weight these clinical cues.

The three evidence types $\mathcal E_f$, $\mathcal E_k$, and $\mathcal E_c$ comprise graph paths and retrieval confidence annotations, denoted $(p_f,C_f)$, $(p_k,C_k)$, and $(p_c,C_c)$, respectively.

At round $t$, current findings and the preceding diagnostic state define support ($\rho=+$) and refutation ($\rho=-$) queries for the leading hypotheses:
\begin{equation}
\begin{aligned}
q_t^{y,\rho}&=(T_{\mathrm{find}}^{(t)},T_{\mathrm{clin}},y,\rho,\Delta_{t-1}),\\
\mathcal Q_t&=\{q_t^{y,\rho}:y\in\mathcal Y_{t-1}^{(2)},\ \rho\in\{+,-\}\}.
\end{aligned}
\label{eq:differential_queries}
\end{equation}
For each query, the top-$k$ retriever returns three evidence types:
\begin{equation}
\mathcal E_t^{y,\rho}=\operatorname{Ret}_{\mathrm{KC}}(\mathcal G,q_t^{y,\rho};k)=(\mathcal E_{f,t}^{y,\rho},\mathcal E_{k,t}^{y,\rho},\mathcal E_{c,t}^{y,\rho}).
\label{eq:retrieved_evidence}
\end{equation}
Evidence from the four queries is aggregated by type into $\mathcal E_t=(\mathcal E_{f,t},\mathcal E_{k,t},\mathcal E_{c,t})$, preserving the target disease, query polarity, graph path, retrieval score, and source of each item.

\subsection{Evidence-Grounded Hypothesis Verification}
\label{sec:method_verification}

Retrieved knowledge completes the accumulated evidence for round $t$:
\begin{equation}
\mathcal L_t=\operatorname{Append}
\left(\mathcal L_t^{\mathrm{img}},\mathcal E_t\right).
\label{eq:ledger_update}
\end{equation}
Diagnostic support is assessed by comparing patient findings with retrieved disease associations. Findings are linked to the source slices, and retrieved associations to their graph paths.

The MLLM assesses competing diagnoses using the global rendering, current local CT images, prior state, and accumulated evidence:
\begin{equation}
(\mathbf d_t,\mathcal P_{t+1})=\mathcal M_{\theta}^{\mathrm{ver}}\bigl(I^{\mathrm g},I_t^{\mathrm{loc}},\mathbf c,\mathbf d_{t-1},\mathcal L_t\bigr).
\label{eq:verification}
\end{equation}
Verification revises $\pi_t$, $\mathbf s_t$, and $\Delta_t$. The updated associations $\mathcal B_t$ link each candidate diagnosis to supporting or refuting records in $\mathcal L_t$. The revised ranking defines the leading hypotheses $\mathcal Y_t^{(2)}$ for subsequent CT inspection and evidence retrieval.

Following~\citep{li_dcldsrag_miccai}, verification integrates imaging findings with clinical context to assess support and disease-specific counterevidence for competing diagnoses, with priority given to CT evidence. Unresolved diagnostic distinctions guide subsequent CT inspection, with slice proposals $\mathcal P_{t+1}$ informed by accumulated observations and retrieved knowledge.

\subsection{Diagnostic Acceptance and Termination}
\label{sec:method_stopping}

The accumulated evidence must contain a local CT observation supporting the leading diagnosis ($g_t^{\mathrm{img}}=1$). It must also contain counterevidence for this diagnosis or evidence supporting or refuting the second-ranked diagnosis ($g_t^{\mathrm{diff}}=1$). The conflict indicator $g_t^{\mathrm{conf}}$ equals one when the leading diagnosis has counterevidence and the second-ranked diagnosis has support but no counterevidence.

The two leading diagnoses must remain unchanged in both identity and order between consecutive assessments ($g_t^{\mathrm{stab}}=1$). At $t=1$, the comparison uses the initial ranking $\pi_0$. These binary indicators define the acceptance decision:
\begin{equation}
a_t=g_t^{\mathrm{img}}\,g_t^{\mathrm{diff}}\,(1-g_t^{\mathrm{conf}})\,g_t^{\mathrm{stab}}.
\label{eq:acceptance}
\end{equation}

Verification stops at acceptance or after $T_{\max}$ rounds, with $T_{\max}=2$ in the main experiments. The stopping round and final diagnosis are
\begin{equation}
\begin{aligned}
t^{\star}&=\min\!\left(\{t\in\{1,\ldots,T_{\max}\}:a_t=1\}
\cup\{T_{\max}\}\right),\\
\widehat y&=\pi_{t^{\star}}(1).
\end{aligned}
\label{eq:final_output}
\end{equation}
The final diagnosis is reported with the accumulated evidence and acceptance status. Predictions with $a_{t^\star}=0$ are uncertain.

\begin{algorithm}[!tbp]
\caption{Evidence-Grounded DCLDs-EIR Inference}
\label{alg:dclds_rag}
\begingroup
\small
\algrenewcommand{\algorithmicindent}{1.0em}
\algrenewcommand{\alglinenumber}[1]{\scriptsize #1:}
\algrenewcommand{\algorithmicrequire}{\textbf{Input:}}
\algrenewcommand{\algorithmicensure}{\textbf{Output:}}
\begin{algorithmic}[1]
\setlength{\itemsep}{1.2pt}
\setlength{\parsep}{0pt}
\Require CT volume $X$; clinical context $\mathbf c$;
segmentation model $\mathcal F_\phi$; MLLM $\mathcal M_\theta$; graph $\mathcal G$.
\Require Slice budget $B$; retrieval setting $k$; round limit $T_{\max}$.
\Ensure Diagnosis $\widehat y$, diagnostic state $\mathbf d_{t^\star}$,
acceptance flag $a_{t^\star}$, and accumulated evidence $\mathcal L_{t^\star}$.
\State $(M_{\mathrm{lung}},M_{\mathrm{cyst}})\gets\mathcal F_\phi(X)$
\State Obtain $L$ and $I^{\mathrm g}$ using Eqs.~\eqref{eq:segmentation} and~\eqref{eq:multiview}.
\State $(\mathbf d_0,\mathcal O_0,\mathcal P_1)\gets\mathcal M_\theta^{\mathrm{init}}(I^{\mathrm g},\mathbf c,D)$
\State Derive $\mathcal Y_0^{(2)}$ using Eq.~\eqref{eq:leading_pair}.
\State Construct $\mathcal I$ from the masks and $\mathcal Y_0^{(2)}$ as in Sec.~\ref{sec:method_local}.
\State Initialize $\mathcal L_0$ from $I^{\mathrm g}$, $\mathbf c$, and $\mathcal O_0$.
\State $\mathcal O_{\leq0}\gets\mathcal O_0$; $\mathcal J_0\gets\varnothing$
\State $T_{\mathrm{clin}}\gets$ clinical information in $\mathbf c$
\For{$t=1,\ldots,T_{\max}$}
    \State $\mathcal S_t\gets\operatorname{Select}(\mathcal I,\mathcal P_t,\mathcal J_{t-1};B)$
    \State $I_t^{\mathrm{loc}}\gets\{\mathcal W(X[z,:,:]):z\in\mathcal S_t\}$
    \State $\mathcal O_t\gets\mathcal M_\theta^{\mathrm{obs}}(I_t^{\mathrm{loc}},\mathbf c,\mathbf d_{t-1})$
    \State $\mathcal O_{\leq t}\gets\operatorname{Append}(\mathcal O_{\leq t-1},\mathcal O_t)$
    \State $\mathcal L_t^{\mathrm{img}}\gets\operatorname{Append}(\mathcal L_{t-1},\mathcal O_t)$
    \State $\mathcal J_t\gets\mathcal J_{t-1}\cup\mathcal S_t$
    \State Form $T_{\mathrm{find}}^{(t)}$ from supplied findings and $\mathcal O_{\leq t}$.
    \State Construct $\mathcal Q_t$ using Eq.~\eqref{eq:differential_queries}.
    \State $\mathcal E_t\gets(\varnothing,\varnothing,\varnothing)$
    \ForAll{$q_t^{y,\rho}\in\mathcal Q_t$}
        \State $\mathcal E_t^{y,\rho}\gets\operatorname{Ret}_{\mathrm{KC}}(\mathcal G,q_t^{y,\rho};k)$
        \State Merge $\mathcal E_t^{y,\rho}$ into $\mathcal E_t$ by evidence type.
    \EndFor
    \State $\mathcal L_t\gets\operatorname{Append}(\mathcal L_t^{\mathrm{img}},\mathcal E_t)$
    \State $(\mathbf d_t,\mathcal P_{t+1})\gets\mathcal M_\theta^{\mathrm{ver}}$
    \Statex \hspace{\algorithmicindent}$\bigl(I^{\mathrm g},I_t^{\mathrm{loc}},\mathbf c,\mathbf d_{t-1},\mathcal L_t\bigr)$
    \State Derive $\mathcal Y_t^{(2)}$ using Eq.~\eqref{eq:leading_pair}.
    \State $a_t\gets\operatorname{EvidenceCheck}(\mathbf d_{t-1},\mathbf d_t,\mathcal L_t;t)$
    \Statex \hspace{\algorithmicindent}\textit{Evaluate Eq.~\eqref{eq:acceptance} as described in Sec.~\ref{sec:method_stopping}.}
    \State $t^\star\gets t$
    \If{$a_t=1$}
        \State \textbf{break}
    \EndIf
\EndFor
\State \Return $(\widehat y,\mathbf d_{t^\star},a_{t^\star},\mathcal L_{t^\star})$, with $\widehat y$ from Eq.~\eqref{eq:final_output}.
\end{algorithmic}
\endgroup
\end{algorithm}

\section{Experiments}
\label{sec:experiments}

\subsection{Cohorts and Evaluation Metrics}
\label{subsec:datasets_metrics}

The study comprised three nonoverlapping cohorts totaling 224 CT examinations. The training/reference and internal test cohorts included 78 and 66 examinations, respectively, from the First Affiliated Hospital of USTC. The external test cohort comprised 80 examinations from the Second Xiangya Hospital of Central South University. Reference diagnoses were established using histopathological or guideline-based criteria~\citep{li_dcldsrag_miccai,jia_selfverification_miccai}. Patient-derived retrieval resources and prompt demonstrations were drawn exclusively from the reference cohort. Table~\ref{tab:dclds_cohort} summarizes cohort characteristics.

Diagnostic performance was assessed using four-class accuracy and macro-averaged precision, recall, specificity, and F1 score. Classwise metrics were computed in a one-versus-rest manner. All test examinations were included irrespective of diagnostic acceptance. Bootstrap standard deviations and percentile 95\% confidence intervals were estimated from 5,000 disease-stratified resamples. Paired comparisons used identical resampled examinations across methods.

\begin{table}[pos=!tbp]
\centering
\caption{Cohorts characteristics. Q1 and Q3 denote the first and third quartiles.}
\label{tab:dclds_cohort}
\begingroup
\small
\definecolor{DCLcohortGroup}{gray}{0.95}
\setlength{\tabcolsep}{2pt}
\renewcommand{\arraystretch}{1.12}
\renewcommand{\tabularxcolumn}[1]{m{#1}}
\def\DCLcohortStat#1#2{\shortstack[c]{#1\\{[#2]}}}
\def\DCLcohortLabel#1{\makebox[\linewidth][c]{\shortstack[c]{#1}}}
\begin{tabularx}{\columnwidth}{@{}>{\raggedright\arraybackslash}m{0.3\columnwidth}*{3}{>{\centering\arraybackslash}X}@{}}
\toprule
\multirow[c]{3}{\linewidth}[-1.5ex]{\centering\textbf{Characteristic}}
  & \multicolumn{2}{c}{\textbf{Internal cohort}}
  & \shortstack{\textbf{External}\\\textbf{cohort}} \\
\cmidrule(lr){2-3}\cmidrule(l){4-4}
  & \shortstack{Training /\\retrieval} & Internal test & External test \\
  & ($n=78$) & ($n=66$) & ($n=80$) \\
\midrule
\rowcolor{DCLcohortGroup}[0pt][0pt]
\multicolumn{4}{@{}l@{}}{\textbf{Disease distribution, $n$ (\%)}} \\
\addlinespace[2pt]
\quad BHD & 40 (51.3) & 34 (51.5) & 14 (17.5) \\
\quad LAM & 14 (17.9) & 12 (18.2) & 33 (41.3) \\
\quad PLCH & 6 (7.7) & 2 (3.0) & 15 (18.8) \\
\quad LIP & 18 (23.1) & 18 (27.3) & 18 (22.5) \\
\addlinespace[4pt]
\rowcolor{DCLcohortGroup}[0pt][0pt]
\multicolumn{4}{@{}l@{}}{\textbf{CT volume characteristics, median [Q1, Q3]}} \\
\addlinespace[2pt]
\DCLcohortLabel{Voxel spacing\\$x/y$ axes (mm)} & \DCLcohortStat{0.78}{0.73, 0.78} & \DCLcohortStat{0.74}{0.70, 0.78} & \DCLcohortStat{0.66}{0.63, 0.69} \\
\DCLcohortLabel{Range} & 0.57--0.97 & 0.39--0.94 & 0.54--0.86 \\
\addlinespace[2pt]
\DCLcohortLabel{Voxel spacing\\$z$ axis (mm)} & \DCLcohortStat{1.25}{1.25, 1.25} & \DCLcohortStat{1.25}{1.00, 1.25} & \DCLcohortStat{0.90}{0.70, 1.00} \\
\DCLcohortLabel{Range} & 0.40--6.00 & 0.62--5.00 & 0.63--5.00 \\
\addlinespace[2pt]
\DCLcohortLabel{Volume depth\\(voxels)} & \DCLcohortStat{279.0}{260.0, 305.0} & \DCLcohortStat{293.0}{265.0, 330.8} & \DCLcohortStat{363.0}{321.8, 438.5} \\
\DCLcohortLabel{Range} & 58--828 & 66--606 & 59--593 \\
\bottomrule
\end{tabularx}
\endgroup
\end{table}

\begin{figure}[pos=!tbp]
\centering
\includegraphics[width=\columnwidth]{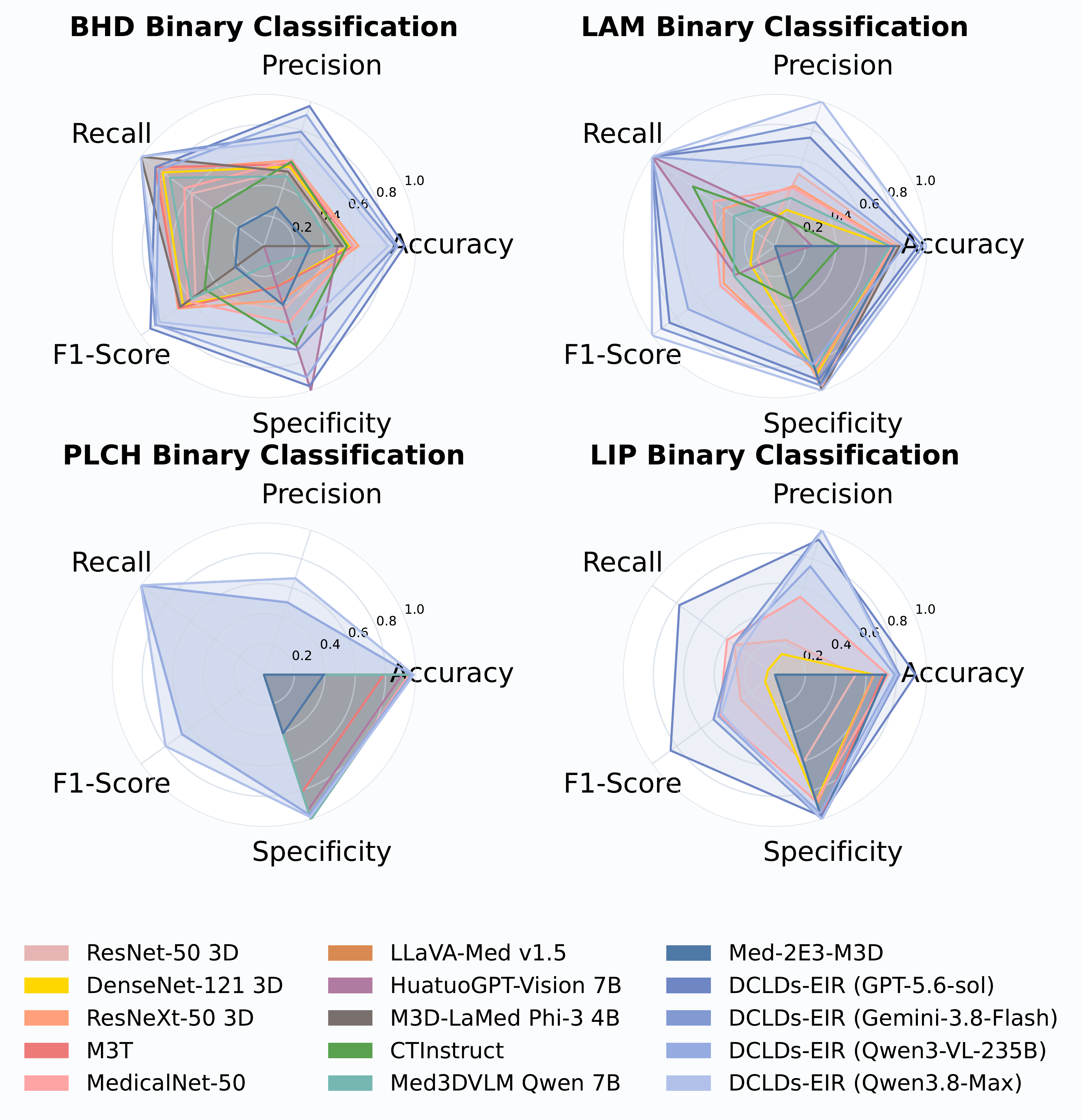}
\caption{Disease-specific binary classification results on the external cohort. The four radar plots compare across conventional 3D classifiers, medical MLLMs, and DCLDs-EIR with different MLLMs.}
\label{fig:dclds_classwise}
\end{figure}

\begingroup
\definecolor{DCLcmpBest}{RGB}{244,204,204}
\definecolor{DCLcmpSecond}{RGB}{255,242,204}
\definecolor{DCLcmpThird}{RGB}{201,218,248}
\definecolor{groupcolor}{RGB}{242,242,242}

\def\DCLval#1#2{#1\ensuremath{_{\scriptstyle\pm #2}}}
\def\DCLstat#1#2#3{\ensuremath{\vcenter{\offinterlineskip
  \halign{\hfil##\hfil\cr
    \DCLval{#1}{#2}\cr
    \noalign{\kern1.5pt}
    {\fontsize{6}{7}\selectfont\thinmuskip=1mu\ensuremath{#3}}\cr
  }}}}

\begin{table*}[pos=!tbp]
\centering
\caption{Four-class DCLDs diagnostic performance (\%) of 3D methods on the internal
and external test cohorts. Point estimates $\pm$ bootstrap SD are
shown above percentile 95\% CIs (Metrics except accuracy are
macro-averaged). Red, yellow, and blue mark the top three estimates
per column. Bold and underlining indicate the first two ranks.}
\label{tab:dclds_comparison_3d}

\fontsize{8}{9.5}\selectfont
\setlength{\tabcolsep}{0.5pt}
\renewcommand{\arraystretch}{1.05}
\renewcommand{\tabularxcolumn}[1]{m{#1}}

\begin{tabularx}{\textwidth}
{@{}>{\raggedright\arraybackslash}m{0.17\textwidth}
*{10}{>{\centering\arraybackslash}X}@{}}
\toprule
\multirow[c]{2}{\linewidth}[-0.5ex]
{\centering\shortstack[c]{Method}}
& \multicolumn{5}{c}{Internal test}
& \multicolumn{5}{c}{External test} \\
\cmidrule(lr){2-6}\cmidrule(lr){7-11}
& \mbox{ACC\,$\uparrow$}
& \mbox{Prec.\,$\uparrow$}
& \mbox{Rec.\,$\uparrow$}
& \mbox{Spec.\,$\uparrow$}
& \mbox{F1\,$\uparrow$}
& \mbox{ACC\,$\uparrow$}
& \mbox{Prec.\,$\uparrow$}
& \mbox{Rec.\,$\uparrow$}
& \mbox{Spec.\,$\uparrow$}
& \mbox{F1\,$\uparrow$} \\
\midrule

\rowcolor{groupcolor}
\multicolumn{11}{l}{\textbf{3D classifiers}} \\
\addlinespace[2pt]

ResNet-50 3D~\citep{resnet}
& \DCLstat{40.91}{5.55}{[30.30,\,51.52]}
& \DCLstat{31.66}{10.40}{[14.74,\,47.94]}
& \DCLstat{25.12}{4.08}{[17.32,\,33.46]}
& \DCLstat{75.19}{2.29}{[70.75,\,79.67]}
& \DCLstat{24.44}{4.49}{[15.88,\,33.22]}
& \DCLstat{28.75}{3.46}{[22.50,\,35.00]}
& \DCLstat{38.81}{11.86}{[13.63,\,60.84]}
& \DCLstat{32.55}{4.09}{[24.77,\,40.72]}
& \DCLstat{77.17}{1.11}{[75.06,\,79.34]}
& \DCLstat{22.95}{4.52}{[14.37,\,31.80]} \\
\addlinespace[5pt]

DenseNet-121 3D~\citep{densenet}
& \DCLstat{46.97}{4.15}{[37.88,\,54.55]}
& \DCLstat{23.55}{5.59}{[13.46,\,35.43]}
& \DCLstat{26.14}{3.40}{[19.85,\,33.17]}
& \DCLstat{76.13}{1.65}{[72.95,\,79.46]}
& \DCLstat{23.47}{3.88}{[16.25,\,31.54]}
& \DCLstat{28.75}{4.68}{[20.00,\,37.50]}
& \DCLstat{24.41}{4.04}{[16.72,\,32.36]}
& \DCLstat{29.27}{4.82}{[19.99,\,38.57]}
& \DCLstat{76.67}{1.60}{[73.56,\,79.82]}
& \DCLstat{23.79}{3.86}{[16.36,\,31.07]} \\
\addlinespace[5pt]

ResNeXt-50 3D~\citep{resnext}
& \DCLstat{51.52}{4.00}{[43.94,\,59.09]}
& \DCLstat{25.21}{3.30}{[18.82,\,32.05]}
& \DCLstat{31.74}{3.83}{[24.14,\,39.34]}
& \DCLstat{78.53}{1.63}{[75.20,\,81.66]}
& \DCLstat{27.89}{3.39}{[20.83,\,34.30]}
& \DCLstat{37.50}{4.74}{[28.75,\,46.25]}
& \DCLstat{27.61}{4.51}{[18.84,\,36.73]}
& \DCLstat{34.51}{4.59}{[25.61,\,43.40]}
& \DCLstat{77.83}{1.70}{[74.55,\,81.16]}
& \DCLstat{28.82}{4.16}{[20.73,\,36.78]} \\
\addlinespace[5pt]

M3T~\citep{m3t}
& \DCLstat{45.45}{2.87}{[39.39,\,50.00]}
& \DCLstat{14.15}{0.79}{[12.73,\,15.82]}
& \DCLstat{22.06}{1.39}{[19.12,\,24.26]}
& \DCLstat{76.82}{1.28}{[74.35,\,79.30]}
& \DCLstat{17.24}{0.88}{[15.43,\,18.89]}
& \DCLstat{20.00}{3.14}{[13.75,\,26.25]}
& \DCLstat{17.83}{6.98}{[7.18,\,34.91]}
& \DCLstat{27.70}{4.21}{[19.40,\,36.03]}
& \DCLstat{75.63}{0.96}{[73.72,\,77.55]}
& \DCLstat{16.41}{3.73}{[9.42,\,23.93]} \\
\addlinespace[5pt]

MedicalNet~\citep{medicalnet}
& \DCLstat{53.03}{5.81}{[42.42,\,63.64]}
& \DCLstat{38.33}{5.07}{[28.53,\,48.56]}
& \DCLstat{38.40}{5.05}{[28.55,\,48.28]}
& \DCLstat{80.60}{2.40}{[75.91,\,85.24]}
& \DCLstat{37.89}{4.93}{[27.85,\,47.40]}
& \DCLstat{22.50}{4.41}{[13.75,\,31.25]}
& \DCLstat{21.44}{4.30}{[13.47,\,30.09]}
& \DCLstat{24.22}{4.74}{[14.97,\,33.71]}
& \DCLstat{74.57}{1.53}{[71.65,\,77.64]}
& \DCLstat{20.19}{3.87}{[12.75,\,27.80]} \\
\addlinespace[5pt]

\rowcolor{groupcolor}
\multicolumn{11}{l}{\textbf{3D medical MLLMs}} \\
\addlinespace[2pt]

M3D-LaMed Phi-3~\citep{p1577_r08}
& \DCLstat{51.52}{0.00}{[51.52,\,51.52]}
& \DCLstat{12.88}{0.00}{[12.88,\,12.88]}
& \DCLstat{25.00}{0.00}{[25.00,\,25.00]}
& \DCLstat{75.00}{0.00}{[75.00,\,75.00]}
& \DCLstat{17.00}{0.00}{[17.00,\,17.00]}
& \DCLstat{17.50}{0.00}{[17.50,\,17.50]}
& \DCLstat{4.37}{0.00}{[4.37,\,4.37]}
& \DCLstat{25.00}{0.00}{[25.00,\,25.00]}
& \DCLstat{75.00}{0.00}{[75.00,\,75.00]}
& \DCLstat{7.45}{0.00}{[7.45,\,7.45]} \\
\addlinespace[5pt]

CTInstruct~\citep{ctinstruct}
& \DCLstat{33.33}{5.00}{[24.24,\,43.94]}
& \DCLstat{19.35}{2.67}{[14.06,\,24.71]}
& \DCLstat{26.96}{4.08}{[18.63,\,34.80]}
& \DCLstat{76.45}{1.98}{[72.57,\,80.35]}
& \DCLstat{19.48}{2.83}{[13.92,\,24.96]}
& \DCLstat{43.75}{3.08}{[37.50,\,50.00]}
& \DCLstat{19.87}{2.93}{[14.29,\,25.71]}
& \DCLstat{31.66}{3.48}{[25.05,\,38.53]}
& \DCLstat{77.60}{1.21}{[75.24,\,79.95]}
& \DCLstat{23.93}{3.12}{[17.73,\,29.83]} \\
\addlinespace[5pt]

Med3DVLM~\citep{med3dvlm}
& \DCLstat{45.45}{4.49}{[36.36,\,54.55]}
& \DCLstat{20.37}{3.70}{[13.18,\,28.01]}
& \DCLstat{27.45}{3.89}{[19.73,\,35.17]}
& \DCLstat{74.42}{1.72}{[71.01,\,77.69]}
& \DCLstat{23.11}{3.67}{[15.66,\,30.09]}
& \DCLstat{17.50}{2.77}{[12.50,\,22.50]}
& \DCLstat{9.52}{2.76}{[4.10,\,15.26]}
& \DCLstat{21.92}{2.98}{[15.80,\,27.27]}
& \DCLstat{73.32}{1.08}{[71.11,\,75.37]}
& \DCLstat{10.07}{1.96}{[6.40,\,14.14]} \\
\addlinespace[5pt]

Med-2E3-M3D~\citep{med2e3}
& \DCLstat{10.61}{3.59}{[4.55,\,18.18]}
& \DCLstat{6.73}{1.83}{[3.00,\,10.16]}
& \DCLstat{5.15}{1.74}{[2.21,\,8.82]}
& \DCLstat{69.39}{1.36}{[66.72,\,72.05]}
& \DCLstat{5.83}{1.76}{[2.54,\,9.17]}
& \DCLstat{22.50}{1.93}{[18.75,\,26.25]}
& \DCLstat{17.57}{5.20}{[4.93,\,29.81]}
& \DCLstat{30.36}{2.75}{[25.00,\,35.71]}
& \DCLstat{76.17}{0.59}{[75.02,\,77.33]}
& \DCLstat{15.93}{3.44}{[8.24,\,22.26]} \\
\addlinespace[5pt]

\rowcolor{groupcolor}
\multicolumn{11}{l}{\textbf{3D DCLDs-EIR}} \\
\addlinespace[2pt]

\textit{w/} Gemini 3.8 Flash~\citep{gemini38flash}
& \cellcolor{DCLcmpSecond}\DCLstat{\underline{81.82}}{3.04}{[75.76,\,87.88]}
& \cellcolor{DCLcmpSecond}\DCLstat{\underline{82.86}}{5.43}{[75.73,\,93.81]}
& \cellcolor{DCLcmpSecond}\DCLstat{\underline{83.33}}{2.79}{[77.78,\,88.89]}
& \cellcolor{DCLcmpSecond}\DCLstat{\underline{91.65}}{1.49}{[88.92,\,94.63]}
& \cellcolor{DCLcmpSecond}\DCLstat{\underline{77.65}}{5.62}{[66.95,\,88.38]}
& \cellcolor{DCLcmpThird}\DCLstat{83.75}{3.99}{[75.00,\,91.25]}
& \cellcolor{DCLcmpThird}\DCLstat{85.18}{3.04}{[79.39,\,91.00]}
& \cellcolor{DCLcmpThird}\DCLstat{83.64}{4.10}{[75.23,\,91.26]}
& \cellcolor{DCLcmpThird}\DCLstat{94.87}{1.26}{[92.22,\,97.20]}
& \cellcolor{DCLcmpThird}\DCLstat{82.41}{4.27}{[73.44,\,90.16]} \\
\addlinespace[5pt]

\textit{w/} Qwen3.8-Max~\citep{qwen38max}
& \cellcolor{DCLcmpThird}\DCLstat{80.30}{2.85}{[74.24,\,86.36]}
& \cellcolor{DCLcmpBest}\DCLstat{\textbf{85.14}}{5.26}{[78.48,\,94.32]}
& \cellcolor{DCLcmpThird}\DCLstat{81.94}{2.61}{[76.39,\,87.50]}
& \cellcolor{DCLcmpThird}\DCLstat{90.23}{1.46}{[87.50,\,93.36]}
& \cellcolor{DCLcmpThird}\DCLstat{77.12}{5.40}{[67.40,\,87.18]}
& \cellcolor{DCLcmpSecond}\DCLstat{\underline{86.25}}{3.64}{[78.75,\,92.50]}
& \cellcolor{DCLcmpSecond}\DCLstat{\underline{88.32}}{2.14}{[84.44,\,92.50]}
& \cellcolor{DCLcmpSecond}\DCLstat{\underline{85.94}}{3.84}{[78.04,\,93.06]}
& \cellcolor{DCLcmpSecond}\DCLstat{\underline{95.68}}{1.16}{[93.25,\,97.73]}
& \cellcolor{DCLcmpSecond}\DCLstat{\underline{85.14}}{3.79}{[77.39,\,92.13]} \\
\addlinespace[5pt]

\textbf{\textit{w/} GPT-5.6 Sol}~\citep{gpt56sol}
& \cellcolor{DCLcmpBest}\DCLstat{\textbf{87.88}}{3.93}{[80.30,\,95.45]}
& \cellcolor{DCLcmpThird}\DCLstat{78.78}{5.77}{[71.39,\,93.75]}
& \cellcolor{DCLcmpBest}\DCLstat{\textbf{91.50}}{2.85}{[85.87,\,96.49]}
& \cellcolor{DCLcmpBest}\DCLstat{\textbf{96.06}}{1.31}{[93.33,\,98.50]}
& \cellcolor{DCLcmpBest}\DCLstat{\textbf{82.38}}{5.51}{[73.13,\,94.32]}
& \cellcolor{DCLcmpBest}\DCLstat{\textbf{90.00}}{3.35}{[82.50,\,96.25]}
& \cellcolor{DCLcmpBest}\DCLstat{\textbf{89.63}}{3.29}{[82.94,\,95.58]}
& \cellcolor{DCLcmpBest}\DCLstat{\textbf{89.31}}{3.72}{[81.42,\,95.94]}
& \cellcolor{DCLcmpBest}\DCLstat{\textbf{96.82}}{1.06}{[94.48,\,98.81]}
& \cellcolor{DCLcmpBest}\DCLstat{\textbf{89.05}}{3.66}{[81.25,\,95.64]} \\
\addlinespace[5pt]

\bottomrule
\end{tabularx}
\end{table*}
\endgroup

\begingroup
\definecolor{DCLcmpBest}{RGB}{244,204,204}
\definecolor{DCLcmpSecond}{RGB}{255,242,204}
\definecolor{DCLcmpThird}{RGB}{201,218,248}
\definecolor{groupcolor}{RGB}{242,242,242}

\def\DCLval#1#2{#1\ensuremath{_{\scriptstyle\pm #2}}}
\def\DCLstat#1#2#3{\ensuremath{\vcenter{\offinterlineskip
  \halign{\hfil##\hfil\cr
    \DCLval{#1}{#2}\cr
    \noalign{\kern1.5pt}
    {\fontsize{6}{7}\selectfont
     \thinmuskip=1mu\ensuremath{#3}}\cr
  }}}}

\newcommand{\DCLmethod}[2]{%
  #1\\[-1pt]
  \citep{#2}%
}

\begin{table*}[pos=!tbp]
\centering
\caption{Four-class DCLDs diagnostic performance (\%) of 2D methods.}
\label{tab:dclds_comparison_2d}

\fontsize{8}{9.5}\selectfont
\setlength{\tabcolsep}{0.5pt}
\renewcommand{\arraystretch}{1.05}
\renewcommand{\tabularxcolumn}[1]{m{#1}}

\begin{tabularx}{\textwidth}{@{}
  >{\raggedright\arraybackslash}m{0.17\textwidth}
  *{10}{>{\centering\arraybackslash}X}@{}}
\toprule
\multirow[c]{2}{\linewidth}[-0.5ex]{%
  \centering\shortstack[c]{Method}}
 & \multicolumn{5}{c}{Internal test}
 & \multicolumn{5}{c}{External test} \\
\cmidrule(lr){2-6}\cmidrule(lr){7-11}
 & \mbox{ACC\,$\uparrow$}
 & \mbox{Prec.\,$\uparrow$}
 & \mbox{Rec.\,$\uparrow$}
 & \mbox{Spec.\,$\uparrow$}
 & \mbox{F1\,$\uparrow$}
 & \mbox{ACC\,$\uparrow$}
 & \mbox{Prec.\,$\uparrow$}
 & \mbox{Rec.\,$\uparrow$}
 & \mbox{Spec.\,$\uparrow$}
 & \mbox{F1\,$\uparrow$} \\
\midrule

\rowcolor{groupcolor}
\multicolumn{11}{l}{\textbf{2D medical MLLMs}} \\
\addlinespace[2pt]

\DCLmethod{LLaVA-Med v1.5}{p1577_r07}
 & \DCLstat{51.52}{0.00}{[51.52,\,51.52]}
 & \DCLstat{12.88}{0.00}{[12.88,\,12.88]}
 & \DCLstat{25.00}{0.00}{[25.00,\,25.00]}
 & \DCLstat{75.00}{0.00}{[75.00,\,75.00]}
 & \DCLstat{17.00}{0.00}{[17.00,\,17.00]}
 & \DCLstat{17.50}{0.00}{[17.50,\,17.50]}
 & \DCLstat{4.37}{0.00}{[4.37,\,4.37]}
 & \DCLstat{25.00}{0.00}{[25.00,\,25.00]}
 & \DCLstat{75.00}{0.00}{[75.00,\,75.00]}
 & \DCLstat{7.45}{0.00}{[7.45,\,7.45]} \\
\addlinespace[5pt]

\DCLmethod{HuatuoGPT-Vision}{huatuo}
 & \DCLstat{18.18}{0.00}{[18.18,\,18.18]}
 & \DCLstat{4.84}{0.15}{[4.62,\,5.17]}
 & \DCLstat{25.00}{0.00}{[25.00,\,25.00]}
 & \DCLstat{75.29}{0.13}{[75.07,\,75.58]}
 & \DCLstat{8.11}{0.21}{[7.79,\,8.57]}
 & \DCLstat{41.25}{0.00}{[41.25,\,41.25]}
 & \DCLstat{10.31}{0.00}{[10.31,\,10.31]}
 & \DCLstat{25.00}{0.00}{[25.00,\,25.00]}
 & \DCLstat{75.00}{0.00}{[75.00,\,75.00]}
 & \DCLstat{14.60}{0.00}{[14.60,\,14.60]} \\
\addlinespace[5pt]

\rowcolor{groupcolor}
\multicolumn{11}{l}{%
  \textbf{2D general MLLMs}} \\
\addlinespace[2pt]

\DCLmethod{GPT-5.6 Sol}{gpt56sol}
 & \DCLstat{81.82}{3.02}{[75.76,\,87.88]}
 & \cellcolor{DCLcmpBest}
   \DCLstat{\textbf{91.97}}{1.80}{[88.89,\,94.77]}
 & \DCLstat{83.33}{2.77}{[77.78,\,88.89]}
 & \DCLstat{90.94}{1.54}{[88.14,\,93.75]}
 & \DCLstat{83.02}{3.87}{[74.48,\,90.23]}
 & \DCLstat{46.25}{5.58}{[35.00,\,57.50]}
 & \DCLstat{64.13}{8.03}{[45.30,\,76.94]}
 & \DCLstat{54.20}{4.33}{[45.58,\,62.58]}
 & \DCLstat{83.21}{1.62}{[80.07,\,86.38]}
 & \DCLstat{47.14}{5.38}{[36.31,\,57.14]} \\
\addlinespace[5pt]

\DCLmethod{Gemini 3.8 Flash}{gemini38flash}
 & \DCLstat{75.76}{2.02}{[72.73,\,80.30]}
 & \DCLstat{88.09}{8.93}{[60.73,\,92.00]}
 & \DCLstat{77.78}{1.85}{[75.00,\,81.94]}
 & \DCLstat{88.45}{1.10}{[86.57,\,90.80]}
 & \DCLstat{73.21}{3.53}{[66.68,\,79.89]}
 & \DCLstat{42.50}{4.40}{[33.75,\,51.25]}
 & \DCLstat{57.21}{10.83}{[38.30,\,75.10]}
 & \DCLstat{48.85}{4.27}{[40.30,\,57.00]}
 & \DCLstat{82.07}{1.46}{[79.15,\,84.92]}
 & \DCLstat{41.08}{4.54}{[31.92,\,49.66]} \\
\addlinespace[5pt]

\DCLmethod{Qwen3.8-Max}{qwen38max}
 & \DCLstat{68.18}{2.27}{[63.64,\,72.73]}
 & \DCLstat{65.45}{0.42}{[64.66,\,66.35]}
 & \DCLstat{68.75}{3.12}{[62.50,\,75.00]}
 & \DCLstat{83.59}{1.17}{[81.25,\,85.94]}
 & \DCLstat{65.53}{2.44}{[60.14,\,69.77]}
 & \DCLstat{45.00}{4.83}{[36.25,\,55.00]}
 & \cellcolor{DCLcmpBest}
   \DCLstat{\textbf{76.25}}{4.94}{[64.58,\,81.86]}
 & \DCLstat{50.56}{4.66}{[41.87,\,59.80]}
 & \DCLstat{83.69}{1.48}{[81.01,\,86.72]}
 & \DCLstat{47.10}{5.67}{[35.30,\,57.40]} \\
\addlinespace[5pt]

\rowcolor{groupcolor}
\multicolumn{11}{l}{%
  \textbf{2D DCLDs-EIR (DCLDs-RAG)}~\citep{li_dcldsrag_miccai}} \\
\addlinespace[2pt]

\DCLmethod{\textit{w/} Qwen-VL-Max}{qwenvlmax}
 & \DCLstat{80.30}{4.92}{[71.21,\,89.39]}
 & \DCLstat{83.86}{12.03}{[53.40,\,91.26]}
 & \DCLstat{74.02}{9.70}{[56.54,\,91.42]}
 & \DCLstat{92.36}{1.97}{[88.37,\,95.95]}
 & \DCLstat{76.67}{10.22}{[54.52,\,90.28]}
 & \DCLstat{63.75}{4.92}{[53.75,\,72.50]}
 & \DCLstat{61.25}{6.10}{[49.62,\,73.35]}
 & \DCLstat{60.17}{5.28}{[49.77,\,70.23]}
 & \DCLstat{88.17}{1.62}{[84.93,\,91.26]}
 & \DCLstat{58.48}{5.55}{[47.29,\,68.74]} \\
\addlinespace[5pt]

\DCLmethod{\textit{w/} Claude Opus 4.5}{claude45}
 & \cellcolor{DCLcmpThird}
   \DCLstat{87.88}{3.89}{[80.30,\,95.45]}
 & \cellcolor{DCLcmpSecond}
   \DCLstat{\underline{89.82}}{11.81}{[60.15,\,95.75]}
 & \DCLstat{81.05}{9.15}{[65.03,\,96.41]}
 & \cellcolor{DCLcmpThird}
   \DCLstat{95.43}{1.52}{[92.25,\,98.18]}
 & \DCLstat{83.10}{9.85}{[62.09,\,95.46]}
 & \DCLstat{48.75}{5.12}{[38.75,\,58.75]}
 & \DCLstat{56.18}{5.64}{[44.54,\,66.56]}
 & \DCLstat{50.85}{5.14}{[40.71,\,60.71]}
 & \DCLstat{83.10}{1.72}{[79.71,\,86.43]}
 & \DCLstat{47.77}{5.50}{[36.54,\,57.83]} \\
\addlinespace[5pt]

\DCLmethod{\textit{w/} GPT-5.2}{gpt52}
 & \cellcolor{DCLcmpThird}
   \DCLstat{87.88}{3.38}{[81.78,\,93.94]}
 & \DCLstat{84.62}{5.51}{[76.89,\,96.05]}
 & \cellcolor{DCLcmpThird}
   \DCLstat{88.19}{3.42}{[81.25,\,94.44]}
 & \DCLstat{94.78}{1.55}{[91.65,\,97.58]}
 & \cellcolor{DCLcmpThird}
   \DCLstat{84.25}{5.31}{[74.05,\,94.24]}
 & \cellcolor{DCLcmpSecond}
   \DCLstat{\underline{68.75}}{4.63}{[60.00,\,77.50]}
 & \DCLstat{68.00}{5.29}{[57.32,\,78.12]}
 & \cellcolor{DCLcmpThird}
   \DCLstat{66.72}{4.90}{[57.08,\,76.10]}
 & \cellcolor{DCLcmpSecond}
   \DCLstat{\underline{89.91}}{1.51}{[86.89,\,92.75]}
 & \cellcolor{DCLcmpSecond}
   \DCLstat{\underline{64.57}}{5.30}{[53.91,\,74.21]} \\
\addlinespace[5pt]

\DCLmethod{\textit{w/} Gemini 2.5 Pro}{gemini25pro}
 & \cellcolor{DCLcmpSecond}
   \DCLstat{\underline{90.91}}{3.00}{[84.85,\,96.97]}
 & \DCLstat{87.11}{5.48}{[79.66,\,97.97]}
 & \cellcolor{DCLcmpSecond}
   \DCLstat{\underline{91.67}}{2.75}{[86.11,\,97.22]}
 & \cellcolor{DCLcmpSecond}
   \DCLstat{\underline{96.02}}{1.38}{[93.21,\,98.68]}
 & \cellcolor{DCLcmpSecond}
   \DCLstat{\underline{87.61}}{5.06}{[77.91,\,97.17]}
 & \cellcolor{DCLcmpThird}
   \DCLstat{66.25}{4.83}{[56.25,\,75.00]}
 & \cellcolor{DCLcmpThird}
   \DCLstat{68.44}{5.47}{[57.32,\,79.10]}
 & \cellcolor{DCLcmpSecond}
   \DCLstat{\underline{67.31}}{4.78}{[57.73,\,76.44]}
 & \cellcolor{DCLcmpThird}
   \DCLstat{89.07}{1.59}{[85.82,\,92.14]}
 & \cellcolor{DCLcmpThird}
   \DCLstat{64.02}{5.17}{[53.47,\,73.82]} \\
\addlinespace[5pt]

\DCLmethod{\textit{w/} GPT-5.6 Sol}{gpt56sol}
 & \cellcolor{DCLcmpBest}
   \DCLstat{\textbf{95.45}}{2.51}{[89.39,\,100.00]}
 & \cellcolor{DCLcmpThird}
   \DCLstat{89.46}{5.67}{[81.66,\,100.00]}
 & \cellcolor{DCLcmpBest}
   \DCLstat{\textbf{96.49}}{2.00}{[92.32,\,100.00]}
 & \cellcolor{DCLcmpBest}
   \DCLstat{\textbf{98.31}}{0.97}{[96.22,\,100.00]}
 & \cellcolor{DCLcmpBest}
   \DCLstat{\textbf{92.12}}{4.64}{[83.51,\,100.00]}
 & \cellcolor{DCLcmpBest}
   \DCLstat{\textbf{75.00}}{4.46}{[66.25,\,83.75]}
 & \cellcolor{DCLcmpSecond}
   \DCLstat{\underline{74.51}}{4.77}{[65.03,\,83.62]}
 & \cellcolor{DCLcmpBest}
   \DCLstat{\textbf{71.66}}{5.14}{[61.34,\,81.45]}
 & \cellcolor{DCLcmpBest}
   \DCLstat{\textbf{91.99}}{1.45}{[89.10,\,94.82]}
 & \cellcolor{DCLcmpBest}
   \DCLstat{\textbf{70.93}}{5.26}{[60.09,\,80.51]} \\
\addlinespace[5pt]

\bottomrule
\end{tabularx}
\end{table*}
\endgroup

\begin{table}[pos=!tbp]
\centering
\caption{Net error reduction achieved by DCLDs-EIR across MLLMs. Fixed and New denote corrected and introduced errors.}
\label{tab:dclds_paired}
\begingroup
\fontfamily{ptm}\fontsize{9}{10.4}\selectfont
\definecolor{DCLpairedGroup}{gray}{0.95}
\setlength{\tabcolsep}{0.8pt}
\setlength{\arrayrulewidth}{0.3pt}
\renewcommand{\tabularxcolumn}[1]{m{#1}}

\renewcommand{\arraystretch}{1.00}

\def\DCLpairedRowPad{0.8pt}

\def\DCLpairedCIGap{-1.5pt}

\def\DCLpairedFit#1{%
  \begingroup
  \sbox0{#1}%
  \ifdim\wd0>\dimexpr\linewidth-1.2pt\relax
    \resizebox{\dimexpr\linewidth-1.2pt\relax}{!}{\usebox0}%
  \else
    \usebox0%
  \fi
  \endgroup}

\def\DCLpairedBound#1{\DCLpairedBoundAux#1\relax}
\def\DCLpairedBoundAux#1#2\relax{%
  \ifx-#1\textminus#2\else#1#2\fi}

\def\DCLpairedCI#1#2#3{%
  \raisebox{0pt}%
    [\dimexpr\height+\DCLpairedRowPad\relax]%
    [\dimexpr\depth+\DCLpairedRowPad\relax]{%
      \shortstack[c]{%
        \ensuremath{#1}\\[\DCLpairedCIGap]
        \DCLpairedFit{{\fontsize{8}{9}\selectfont
          [\DCLpairedBound{#2},\,\DCLpairedBound{#3}]}}%
      }%
    }%
}

\begin{tabularx}{\columnwidth}{
  @{}>{\hsize=1.90\hsize\linewidth=\hsize
       \raggedright\arraybackslash}X|
  *{2}{>{\hsize=0.70\hsize\linewidth=\hsize
          \centering\arraybackslash}X|}
  *{2}{>{\hsize=0.62\hsize\linewidth=\hsize
          \centering\arraybackslash}X|}
  >{\hsize=1.23\hsize\linewidth=\hsize
    \centering\arraybackslash}X|
  >{\hsize=1.23\hsize\linewidth=\hsize
    \centering\arraybackslash}X@{}
}
\hline
\multirow[c]{2}{\linewidth}{\centering \textit{w/} MLLMs}
  & \multicolumn{2}{c|}{ACC (\%)}
  & \multicolumn{2}{c|}{Errors}
  & \multicolumn{2}{c@{}}{$\Delta$ [95\% CI]} \\
\cline{2-7}
  & Direct & Full & Fixed & New
  & ACC & Macro-F1 \\
\hline

\rowcolor{DCLpairedGroup}
\multicolumn{7}{@{}l@{}}{%
  \rule[-1pt]{0pt}{11pt}\textbf{Internal test}} \\
\hline

\mbox{GPT-5.6 Sol}
 & 81.82 & 87.88 & 9 & 5
 & \DCLpairedCI{+6.06}{-3.03}{15.15}
 & \DCLpairedCI{-0.63}{-12.56}{14.21} \\

\mbox{Gemini 3.8 Flash}
 & 75.76 & 81.82 & 5 & 1
 & \DCLpairedCI{+6.06}{0.00}{12.12}
 & \DCLpairedCI{+4.44}{-6.68}{16.47} \\

\mbox{Qwen3.8-Max}
 & 68.18 & 80.30 & 8 & 0
 & \DCLpairedCI{+12.12}{6.06}{19.70}
 & \DCLpairedCI{+11.59}{0.63}{23.42} \\
\hline

\rowcolor{DCLpairedGroup}
\multicolumn{7}{@{}l@{}}{%
  \rule[-1pt]{0pt}{11pt}\textbf{External test}} \\
\hline

\mbox{GPT-5.6 Sol}
 & 81.25 & 90.00 & 11 & 4
 & \DCLpairedCI{+8.75}{0.00}{17.50}
 & \DCLpairedCI{+9.06}{0.46}{18.72} \\

\mbox{Gemini 3.8 Flash}
 & 72.50 & 83.75 & 10 & 1
 & \DCLpairedCI{+11.25}{3.75}{18.75}
 & \DCLpairedCI{+10.77}{2.95}{19.63} \\

\mbox{Qwen3.8-Max}
 & 45.00 & 86.25 & 34 & 1
 & \DCLpairedCI{+41.25}{31.25}{51.25}
 & \DCLpairedCI{+38.04}{27.62}{49.97} \\
\hline
\end{tabularx}
\endgroup
\end{table}

\begin{figure*}[pos=!tbp]
\centering
\includegraphics[width=\textwidth]{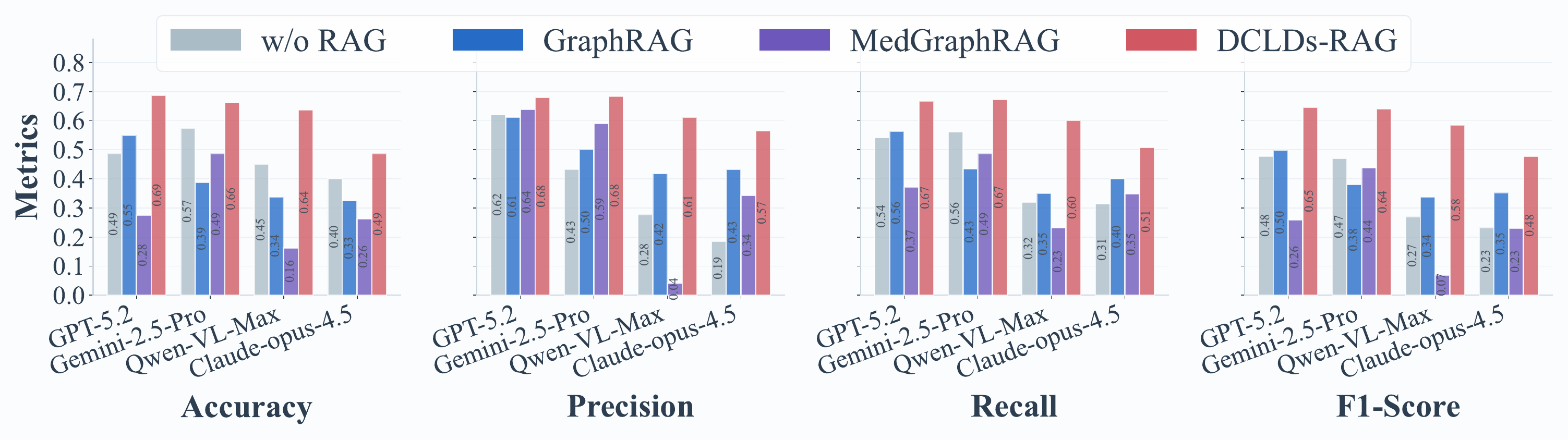}
\caption{External-cohort comparison of RAG methods and our 2D DCLDs-EIR (DCLDs-RAG) setting, using four MLLMs.}
\label{fig:dclds_rag_comparison}
\end{figure*}

\subsection{Implementation Details}
\label{subsec:implementation}

Predicted lung and cyst masks from the segmentation model in~\citep{jia_selfverification_miccai} were used to generate four-view global renderings and guide slice selection. Local inspection used original axial lung-window CT images without segmentation overlays. The per-round slice budget and maximum number of verification rounds were set to $B=4$ and $T_{\max}=2$, respectively. Knowledge and clinical evidence retrieval adopted the graph representation and scoring functions of our previous study~\citep{li_dcldsrag_miccai}, with the graph fixed during inference. GPT-5.6 Sol~\citep{gpt56sol} was used for the main experiments. Gemini 3.8 Flash~\citep{gemini38flash} and Qwen3.8-Max~\citep{qwen38max} were evaluated using the same diagnostic prompts, slice selection rules, and retrieval procedures. All three MLLMs were used without task-specific fine-tuning. OpenAI Codex was used to assist with programming and data-processing tasks during study implementation. The authors reviewed the resulting code and processed outputs and retained responsibility for the data analyses and reported experimental results.

The compared 2D methods used axial CT images and patient information, whereas volumetric 3D baselines received CT volumes. Direct general-purpose MLLMs performed a single diagnostic pass using the supplied images, clinical information, and imaging findings.

\subsection{Quantitative Comparisons}
\label{subsec:main_comparisons}
\label{subsec:paired_analysis}
\label{subsec:efficiency}

DCLDs-EIR achieved the highest external accuracy and macro-F1 among the evaluated methods (Tables~\ref{tab:dclds_comparison_3d} and~\ref{tab:dclds_comparison_2d}). With GPT-5.6 Sol, accuracy reached 90.00\% (95\% CI 82.50--96.25\%) and macro-F1 is 89.05\% (95\% CI 81.25--95.64\%). These results exceeded the earlier 2D retrieval setting with the same MLLM by 15.00 and 18.12 percentage points, respectively. On the internal cohort, DCLDs-EIR achieved 87.88\% accuracy and 82.38\% macro-F1, outperforming the evaluated 3D classifiers and medical MLLMs. The 2D retrieval setting retained higher internal performance, with 95.45\% accuracy and 92.12\% macro-F1, indicating that the advantage of the complete framework was most evident on the external cohort.

Improvements over direct inference extended across all three MLLMs. External accuracy increased by 41.25--43.75 percentage points, accompanied by macro-F1 gains of 38.04--41.91 points. Internally, accuracy improved for all three MLLMs, while macro-F1 increased for Gemini and Qwen. The paired Qwen analysis further demonstrated a substantial net reduction in diagnostic errors (Table~\ref{tab:dclds_paired}). On the external cohort, 34 errors were corrected and one was introduced, yielding an accuracy gain of 41.25 percentage points.

The classwise comparisons in Fig.~\ref{fig:dclds_classwise} show that DCLDs-EIR achieves higher F1 scores across all four diseases than the evaluated conventional classifiers and medical MLLMs, with consistent advantages across MLLM configurations. The improvement is particularly evident for PLCH, where several baselines exhibit high accuracy and specificity but low recall and F1. DCLDs-EIR improves PLCH recall and F1 while maintaining high specificity, demonstrating more effective recognition of positive cases.

The 2D retrieval comparison provides complementary evidence for the disease-specific retrieval design (Fig.~\ref{fig:dclds_rag_comparison}). Across four MLLMs, this approach achieved higher accuracy and macro-F1 than direct inference, GraphRAG~\citep{p323_r12}, and Medical Graph RAG~\citep{p323_r29}. Across four MLLMs, our disease-specific retrieval achieved higher accuracy and F1 than baselines.

The diagnostic gains were accompanied by increased online inference costs (Table~\ref{tab:dclds_efficiency}). With GPT-5.6 Sol, mean inference time increased from 8.29 to 50.37~s per case, and mean cost increased from \$0.0144 to \$0.1844. Gemini achieved the lowest measured time and cost among the framework configurations, at 42.87~s and \$0.0482 per case, while attaining 83.75\% accuracy on the external cohort.

\begin{table}[pos=!tbp]
\centering
\caption{Online inference costs, time and tokens ($10^3$) of DCLDs-EIR across MLLMs. Values are reported as mean $\pm$ SD across cases.}
\label{tab:dclds_efficiency}
\begingroup
\fontsize{9}{10.5}\selectfont
\setlength{\tabcolsep}{0.8pt}
\renewcommand{\arraystretch}{1.05}
\renewcommand{\tabularxcolumn}[1]{m{#1}}

\def\DCLeffval#1#2{%
  \shortstack[c]{%
    \ensuremath{#1}\\[3pt]
    \ensuremath{\pm #2}%
  }%
}

\def\DCLeffmethod#1#2{%
  \shortstack[l]{%
    #1\\[3pt]
    #2%
  }%
}

\begin{tabularx}{\columnwidth}{@{}
  >{\hsize=2.20\hsize\linewidth=\hsize
    \raggedright\arraybackslash}X
  >{\hsize=0.65\hsize\linewidth=\hsize
    \centering\arraybackslash}X
  >{\hsize=0.82\hsize\linewidth=\hsize
    \centering\arraybackslash}X
  >{\hsize=0.75\hsize\linewidth=\hsize
    \centering\arraybackslash}X
  >{\hsize=0.75\hsize\linewidth=\hsize
    \centering\arraybackslash}X
  >{\hsize=0.83\hsize\linewidth=\hsize
    \centering\arraybackslash}X@{}}

\toprule
\makebox[\linewidth][c]{Method}
  & \shortstack{API\\calls}
  & \shortstack{Online\\time (s)}
  & \shortstack{Input}
  & \shortstack{Output}
  & \shortstack{Cost\\(USD)} \\
\midrule

\mbox{GPT-5.6 Sol}
  & \DCLeffval{1.00}{0.00}
  & \DCLeffval{8.29}{1.17}
  & \DCLeffval{2.10}{0.01}
  & \DCLeffval{0.20}{0.04}
  & \DCLeffval{0.0144}{0.0008} \\

\addlinespace[3pt]

\DCLeffmethod{DCLDs-EIR}{\textit{w/} GPT-5.6 Sol}
  & \DCLeffval{3.50}{1.00}
  & \DCLeffval{50.37}{20.20}
  & \DCLeffval{27.76}{12.95}
  & \DCLeffval{2.28}{1.11}
  & \DCLeffval{0.1844}{0.0870} \\

\addlinespace[3pt]

\DCLeffmethod{DCLDs-EIR}{\textit{w/} Gemini 3.8 Flash}
  & \DCLeffval{4.25}{1.50}
  & \DCLeffval{42.87}{13.82}
  & \DCLeffval{52.43}{29.67}
  & \DCLeffval{2.37}{1.19}
  & \DCLeffval{0.0482}{0.0267} \\

\addlinespace[3pt]

\DCLeffmethod{DCLDs-EIR}{\textit{w/} Qwen3.8-Max}
  & \DCLeffval{4.00}{1.15}
  & \DCLeffval{123.63}{57.87}
  & \DCLeffval{32.63}{14.40}
  & \DCLeffval{5.07}{2.51}
  & \DCLeffval{0.0957}{0.0438} \\

\bottomrule
\end{tabularx}
\endgroup
\end{table}

\begingroup
\def\DCLval#1#2{#1\ensuremath{_{\scriptstyle\pm #2}}}
\begin{table}[pos=!tbp]
\centering
\caption{Progressive ablation study of DCLDs-EIR on the internal cohort using GPT-5.6 Sol. Components are added sequentially. KC includes the aligned diagnostic prompt; SV denotes self-verification; MV+DG denotes multiview GSP and diagnosis-guided slice acquisition.}
\label{tab:dclds_ablation}

\fontsize{8}{9.5}\selectfont
\setlength{\tabcolsep}{1.5pt}
\renewcommand{\arraystretch}{1.25}

\begin{tabularx}{\columnwidth}
{@{}l*{5}{>{\centering\arraybackslash}X}@{}}
\toprule
Configuration
& Acc.\,$\uparrow$
& Prec.\,$\uparrow$
& Rec.\,$\uparrow$
& Spec.\,$\uparrow$
& F1\,$\uparrow$ \\
\midrule

GSP
& \DCLval{51.52}{0.00}
& \DCLval{12.88}{0.00}
& \DCLval{25.00}{0.00}
& \DCLval{75.00}{0.00}
& \DCLval{17.00}{0.00} \\

\addlinespace[2pt]
+ Axial
& \DCLval{60.61}{2.52}
& \DCLval{64.17}{2.50}
& \DCLval{58.33}{3.47}
& \DCLval{79.69}{1.30}
& \DCLval{55.59}{4.27} \\

\addlinespace[2pt]
+ KC
& \DCLval{81.82}{4.48}
& \DCLval{74.74}{5.08}
& \DCLval{89.87}{2.68}
& \DCLval{94.20}{1.44}
& \DCLval{78.02}{5.25} \\

\addlinespace[2pt]
+ Iterative SV
& \DCLval{83.33}{4.45}
& \DCLval{72.78}{4.66}
& \DCLval{89.30}{2.96}
& \DCLval{94.97}{1.36}
& \DCLval{76.30}{5.19} \\

\addlinespace[2pt]
+ MV+DG
& \DCLval{\textbf{87.88}}{3.93}
& \DCLval{\textbf{78.78}}{5.77}
& \DCLval{\textbf{91.50}}{2.85}
& \DCLval{\textbf{96.06}}{1.31}
& \DCLval{\textbf{82.38}}{5.51} \\

\bottomrule
\end{tabularx}
\end{table}
\endgroup

\subsection{Ablation and Parameter Analysis}
\label{subsec:component_ablation}
\label{subsec:parameter_sensitivity}

Table~\ref{tab:dclds_ablation} presents a cumulative component ablation on the 66 internal test examinations using GPT-5.6 Sol. Adding axial CT inspection to GSP increases accuracy from 51.52\% to 60.61\% and macro-F1 from 17.00\% to 55.59\%, supporting the contribution of local imaging evidence. Incorporating knowledge and clinical evidence retrieval (KC) together with its aligned diagnostic prompt further increases accuracy and macro-F1 to 81.82\% and 78.02\%, respectively.

Allowing a second verification round increases accuracy to 83.33\% but decreases macro-F1 to 76.30\%. Jointly introducing multiview GSP and diagnosis-guided slice acquisition produces the complete configuration, which achieves 87.88\% accuracy and 82.38\% macro-F1. These gains of 4.55 and 6.08 percentage points over the preceding configuration support combining spatial context and targeted evidence acquisition with iterative verification.

\begin{figure*}[pos=!tbp]
\centering
\includegraphics[width=\textwidth]{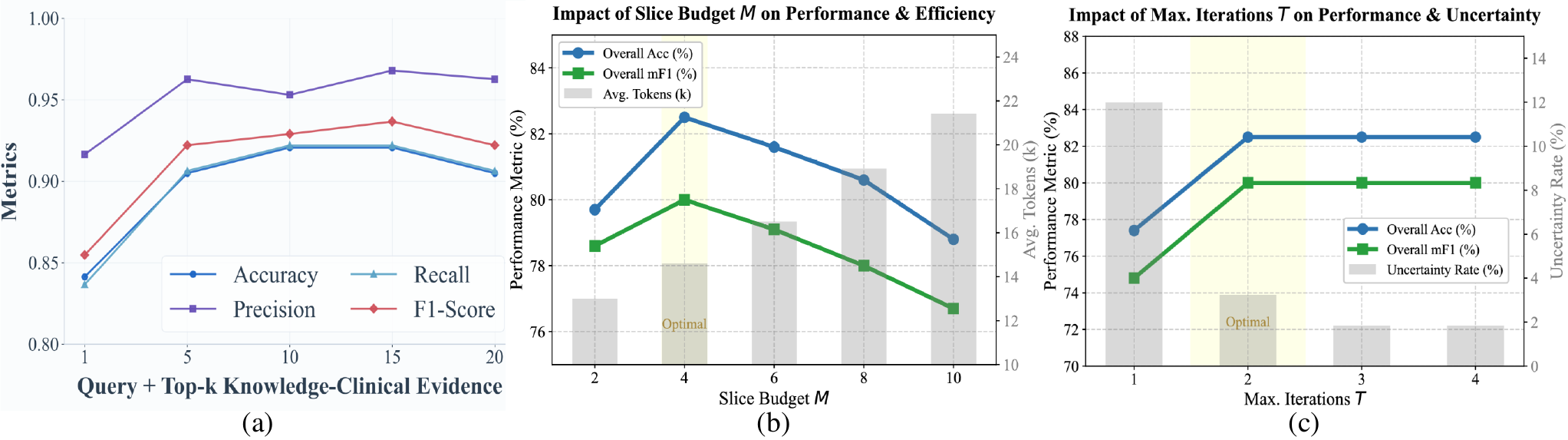}
\caption{(a) Impact of top-k knowledge-clinical envidence on 2D DCLDs-EIR setting. (b) Impact of slice budget M on performance and efficiency, and (c) max iterations T on performance and uncertainty rate on 3D DCLDs-EIR setting \textit{w/o} RAG.}
\label{fig:dclds_parameter_sensitivity}
\end{figure*}

Fig.~\ref{fig:dclds_parameter_sensitivity} summarizes parameter analyses from the retrieval and self-verification. In the 2D setting, accuracy reaches approximately 92\% at $k=10$ and $k=15$, while F1 is highest at $k=15$. Both metrics decrease at $k=20$.
In the self-verification setting, four slices per round yield the highest accuracy and macro-F1, approximately 82.5\% and 80.0\%, respectively, whereas larger budgets increase token consumption and reduce performance. Increasing the round limit from 1 to 2 improves both metrics, with no further gains at three. These component-level trends are consistent with the settings $B=4$ and $T_{\max}=2$ adopted in DCLDs-EIR.

\begin{figure*}[pos=!tbp]
\centering
\includegraphics[width=\textwidth]{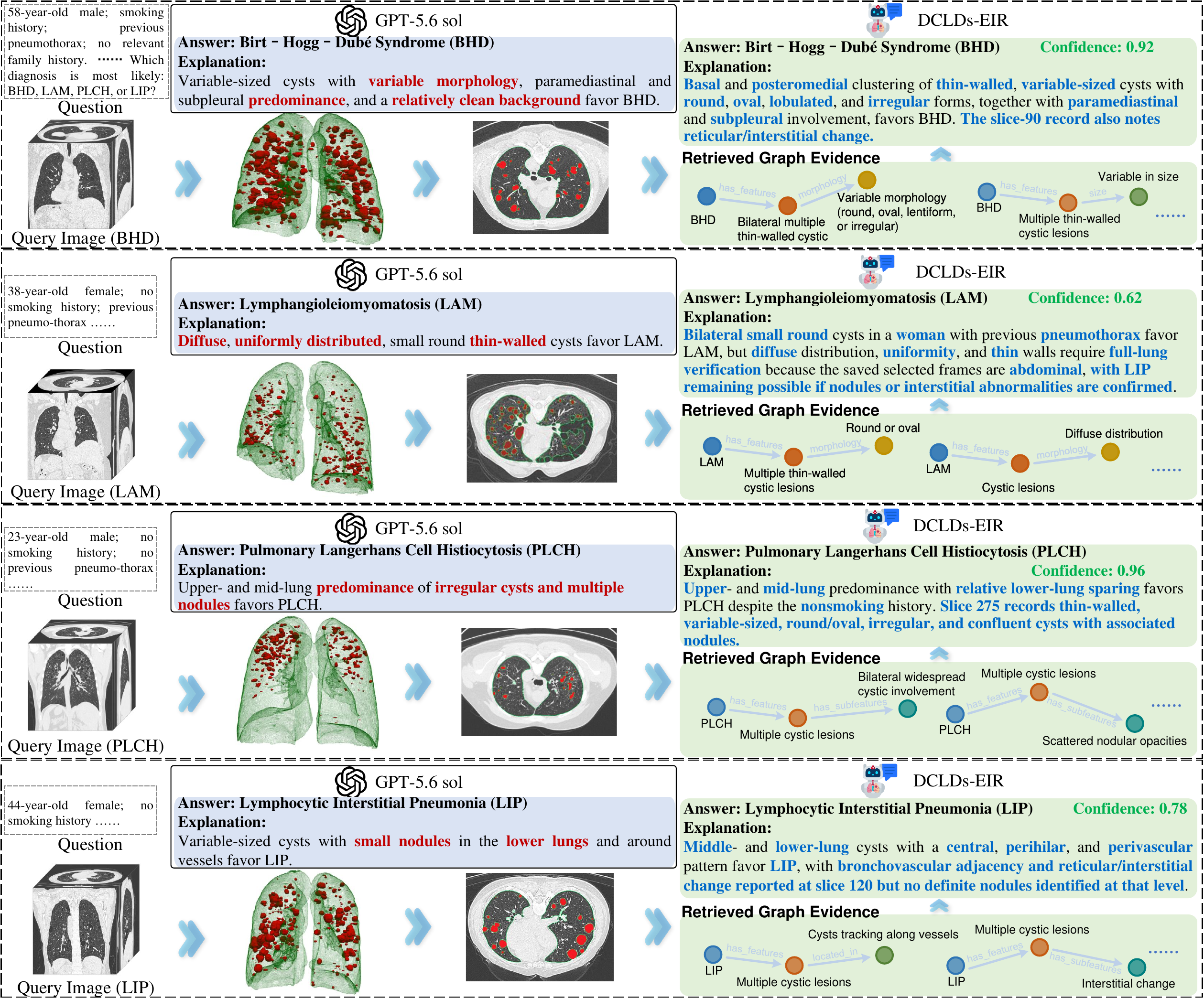}
\caption{Qualitative comparison of direct GPT-5.6 Sol and DCLDs-EIR for four correctly classified cases. Red marks vague or weakly supported statements and blue marks evidence-linked descriptions and explicit uncertainty. Renderings, CT slices, and graph paths show global cyst distributions, local findings, and disease knowledge, respectively.}
\label{fig:dclds_qualitative}
\end{figure*}

\subsection{Qualitative Analysis}
\label{subsec:qualitative}

Fig.~\ref{fig:dclds_qualitative} compares diagnostic explanations for four cases, one per disease, correctly classified by both direct GPT-5.6 Sol inference and DCLDs-EIR. Red denotes vague, incomplete, or insufficiently supported statements, while blue denotes evidence-linked descriptions and explicit uncertainty.

The BHD explanation specifies basal and posteromedial cyst clustering and records reticular changes at slice~90. For LAM, cyst morphology and clinical features are distinguished from distributional characteristics requiring further verification. The PLCH explanation links upper and middle lung predominance with nodules documented at slice~275 while acknowledging the nonsmoking history. The LIP explanation combines perihilar and perivascular cyst distribution with interstitial changes at slice~120 and corresponding graph associations, while limiting the assessment of nodules to that slice. These examples illustrate the localization of supporting observations and explicit qualification of unverified findings.

\section{Discussion}
\label{sec:discussion}

DCLDs-EIR integrates global spatial perception, hypothesis-guided CT inspection, and knowledge--clinical evidence verification into a unified iterative reasoning process. Its main advantage lies in the feedback between imaging evidence acquisition and diagnostic reasoning: whole-lung renderings establish the initial diagnostic context, competing hypotheses guide targeted slice selection, and newly acquired findings refine subsequent knowledge retrieval and hypothesis verification. Importantly, retrieved disease associations are not treated as direct evidence of their presence in individual patients. Instead, supporting and refuting evidence is explicitly assessed against observed imaging findings and clinical information. This coupling provides a more systematic and traceable diagnostic process than relying solely on one-pass MLLM inference or independently retrieved knowledge.

The experimental results support the effectiveness of this design. Ablation studies demonstrate the complementary contributions of global spatial perception, targeted slice inspection, and knowledge--clinical verification. On the independent external cohort of 80 examinations, DCLDs-EIR achieved 90.00\% accuracy and 89.05\% macro-F1 with GPT-5.6 Sol, suggesting its potential to generalize beyond the internal evaluation cohort. The qualitative results further illustrate how linking diagnostic hypotheses to localized CT observations and retrieved graph evidence produces more specific and clinically interpretable explanations.

Nevertheless, performance may still depend on the quality of upstream segmentation and the coverage of the disease knowledge graph. The additional retrieval and verification steps also introduce computational overhead. Future work will investigate more efficient adaptive evidence acquisition, expand the knowledge base, and evaluate the framework on larger prospective multicenter cohorts. Integration into clinician-in-the-loop workflows will also be explored to assess its practical value in rare-disease diagnosis.

\section{Conclusion}
\label{sec:conclusion}

We presented DCLDs-EIR, an evidence-grounded framework integrating multiview global spatial perception, diagnosis-guided slice acquisition, and iterative knowledge--clinical verification for DCLDs diagnosis. By connecting localized CT observations with supporting and refuting graph evidence, the framework refines competing diagnostic hypotheses and provides traceable explanations. Evaluation on internal and external cohorts supports its effectiveness, with the GPT-5.6 Sol implementation achieving 90.00\% accuracy and 89.05\% macro-F1 on the external cohort. The current evidence is limited to retrospective cohorts and limited data. Future work will focus on prospective multicenter validation, improving evidence acquisition, and reducing the computational cost of iterative verification.

\section*{Declaration of generative AI and AI-assisted technologies in the manuscript preparation process}

During the preparation of this work, the authors used OpenAI Codex to assist with programming and data-processing tasks, and ChatGPT (GPT) to support manuscript revision and language polishing. The authors reviewed the code, processed outputs, and AI-assisted manuscript edits as appropriate, and take full responsibility for the data analyses, interpretations, and content of the published article.

\FloatBarrier

\bibliographystyle{cas-model2-names}
\bibliography{main}

\end{document}